\documentclass[10pt,twocolumn,letterpaper]{article}

\usepackage[pagenumbers]{cvpr} 

\definecolor{cvprblue}{rgb}{0.21,0.49,0.74}
\usepackage[pagebackref,breaklinks,colorlinks,allcolors=cvprblue]{hyperref}

\def\paperID{*****} 
\def\confName{CVPR}
\def\confYear{2025}

\title{StyleSSP: \textbf{S}ampling \textbf{S}tart\textbf{P}oint Enhancement for Training-free Diffusion-based Method for Style Transfer}

\author{Ruojun Xu\\
Zhejiang University, Dcar\\
Hangzhou, China\\
{\tt\small ruojunxu@zju.edu.cn}
\and
Weijie Xi\\
Dcar,\\
Hangzhou, China\\
{\tt\small xiweijie@bytedance.com}
\and
XiaoDi Wang\\
Dcar,\\
Beijing,China\\
{\tt\small wangxiaodi.00@bytedance.com}
\and
Yongbo Mao\thanks{Corresponding author}\\
Dcar,\\
Beijing, China\\
{\tt\small maoyongbo@bytedance.com}
\and
Zach Cheng\\
Dcar,\\
Beijing, China\\
{\tt\small chengyi.2024@bytedance.com}
}

\begin{document}
\maketitle
\begin{abstract}
Training-free diffusion-based methods have achieved remarkable success in style transfer, eliminating the need for extensive training or fine-tuning. However, due to the lack of targeted training for style information extraction and constraints on the content image layout, training-free methods often suffer from layout changes of original content and content leakage from style images. Through a series of experiments, we discovered that an effective startpoint in the sampling stage significantly enhances the style transfer process. Based on this discovery, we propose \textbf{StyleSSP}, which focuses on obtaining a better startpoint to address layout changes of original content and content leakage from style image. StyleSSP comprises two key components: (1) \textbf{Frequency Manipulation}: To improve content preservation, we reduce the low-frequency components of the DDIM latent, allowing the sampling stage to pay more attention to the layout of content images; and (2) \textbf{Negative Guidance} via Inversion: To mitigate the content leakage from style image, we employ negative guidance in the inversion stage to ensure that the startpoint of the sampling stage is distanced from the content of style image. Experiments show that StyleSSP surpasses previous training-free style transfer baselines, particularly in preserving original content and minimizing the content leakage from style image.
\end{abstract}
    
\section{Introduction}
\label{sec:intro}

\begin{figure*}[t]
	\centering
	\begin{subfigure}{0.49\linewidth}
		\includegraphics[width=\linewidth]{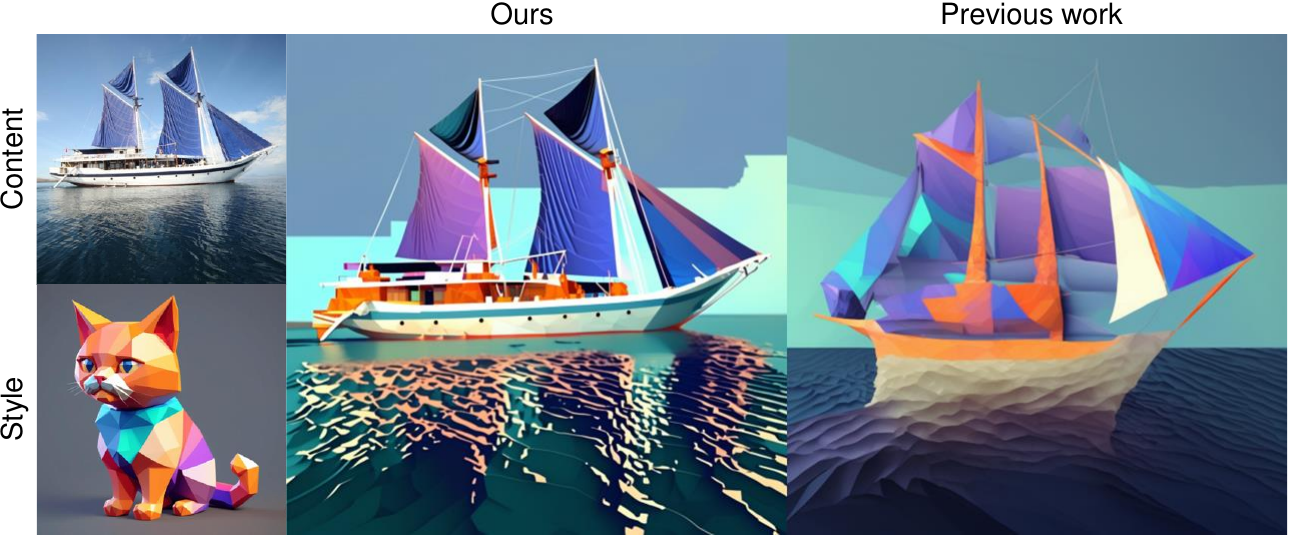}
		\caption{Content Preservation problem.}
		\label{fig:1-a}
	\end{subfigure}
	\begin{subfigure}{0.49\linewidth}
		\includegraphics[width=\linewidth]{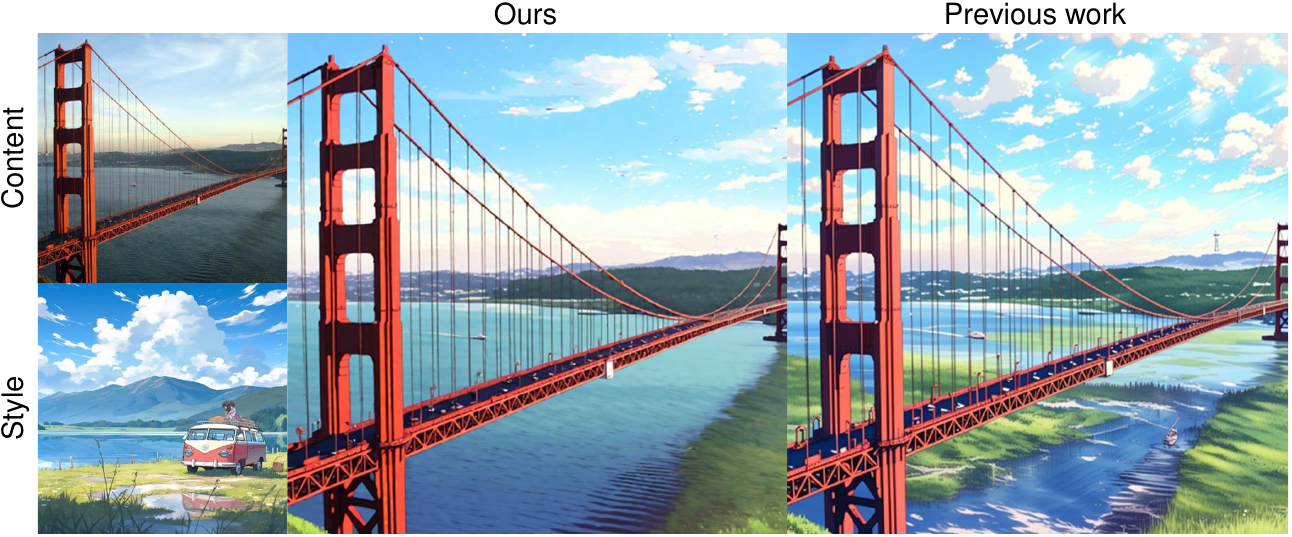}
		\caption{Content Leakage problem.}
		\label{fig:1-b}
	\end{subfigure}
	
	\vspace{0.2cm}
	
	\begin{subfigure}{\linewidth}
		\includegraphics[width=\linewidth]{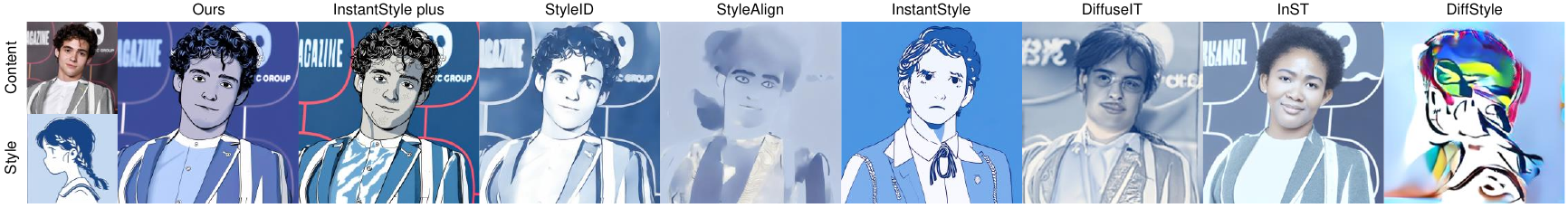}
		\caption{Comparison with other works}
		\label{fig:1-c}
	\end{subfigure}
	\caption{Current problems for style transfer and our improvements. (a) \textbf{Original content changes} in previous work (right) even with ControlNet as an additional content controller. (b) \textbf{Content leakage} from style image in previous work (right), where the river from original image is covered by a lawn that shouldn't exist. (c) Given a style image and content image, StyleSSP is capable of synthesizing new images that achieve the best style transfer effect while preserving the details of original content.}
	\label{fig:1}
\end{figure*}

Recently, Diffusion Models (DMs) have yielded high-quality results in various areas such as text-to-image generation~\cite{T2I_1, T2I_2, T2I_3} and image or video editing~\cite{Edit_1, Edit_2, Edit_3, Edit_4}. As part of image editing, diffusion-based style transfer methods~\cite{StyleID, styleAlign, deadiff, DiffStyle} have garnered widespread attention. These methods enable condition-guided image generation that transfers the style of one image onto another while maintaining the original content.

Previous diffusion-based style transfer methods~\cite{deadiff, InST, MOEController, stylediffusion} leverage the generative capability of pre-trained DMs using inference-stage optimization, yet they are either time-consuming or fail to fully utilize the generative ability of large-scale diffusion models. Based on these challenges, training-free methods~\cite{stylediffusion, StyleID, InstantStyle-Plus, InstantStyle} have been proposed. Although these methods have shown promising results, they still encounter two key issues: (1) \textbf{Content preservation problem}. Due to the lack of constraints directly imposed on the content of generated images during training, training-free methods often struggle to maintain the original semantic and structural content~\cite{NTI}. Although additional modules like ControlNet~\cite{controlnet} can be used as content constraints, experiments have shown that these methods still risk failure (as shown in supplementary materials Sec.~\ref{sec: startpoint effects}). This issue largely arises from the diffusion model’s imbalanced preference for different conditions when multiple conditions are injected into the U-Net during the sampling stage~\cite{brushnet,EMMA}; (2) \textbf{Content leakage from the style image}. Without targeted training for style extraction, training-free methods struggle to effectively decouple the style and content. Therefore, when the style image is directly injected into the pre-trained diffusion model, the generation process is inevitably influenced by the content of style image. Fig.~\ref{fig:1} illustrates examples of these two problems.


To address these challenges, we begin by noting recent advancements in image synthesis tasks with DMs. These studies reveal the significant influence of the initial noise (referred to here as the ``startpoint") on the generated outcome. For example, FreeNoise~\cite{FreeNoise} analyzes the impact of startpoint within video diffusion models, emphasizing the importance of initialization with a sequence of long-range correlated noises. Similarly, FlexiEdit~\cite{flexiedit} enhances the startpoint by reducing low-frequency components, improving the fidelity to editing prompts. While the significance of startpoint selection is increasingly acknowledged in generation and editing tasks, it remains largely unexplored in style transfer. StyleID~\cite{StyleID} does incorporate startpoint manipulation, but only by rescaling the startpoint to offset the pre-trained model’s tendency to generate images with median colors, without fully investigating its role in style transfer.

Inspired by the aforementioned methods, we creatively argue that refining the sampling startpoint is an effective strategy for improving style transfer. Our supplementary materials Sec.~\ref{sec: startpoint effects} further demonstrates this point. In these experiments, we show the startpoint’s significant impact on content preservation and tonal adjustment. Based on these findings, we propose \textbf{StyleSSP} (\textbf{Style} transfer method via \textbf{S}ampling \textbf{S}tart\textbf{P}oint enhancement), a training-free approach that refines the sampling startpoint in diffusion models. To the best of our knowledge, this work is the first to highlight the importance of selecting an effective sampling startpoint to improve style transfer in a training-free, diffusion-based framework.


First, we propose frequency manipulation to improve original content preservation in style transfer. Inspired by FlexiEdit~\cite{flexiedit}, which highlights that high-frequency components are more closely associated with image layout (e.g., contours and details) than low-frequency components, we improve detail preservation by reducing low-frequency components of the DDIM latent, which serves as the sampling startpoint. This refinement enhances the model's ability to retain the image layout during style transfer.

Second, we introduce negative guidance in the DDIM inversion stage to alleviate content leakage from style images. This approach ensures that the sampling startpoint is further ``distanced" from the content of style image. Our experiments (Fig.~\ref{fig:NG}) show that, compared to traditional negative guidance~\cite{oconnor2023stable} applied during the sampling stage, applying guidance in the inversion stage yields superior results by mitigating multi-condition control failures~\cite{brushnet, EMMA}. Additionally, we use the pre-trained IP-Instruct model~\cite{IP-Instruct} as our style and content extractor, providing negative guidance in the inversion stage for a better startpoint.

In summary, our main contributions are as follows:
\begin{itemize}
	\item We propose a novel sampling startpoint enhancement method for training-free diffusion-based style transfer, addressing issues of content leakage from style images and changes in original content. To the best of our knowledge, we are the first work to highlight the importance of the startpoint in this area.
	\item We propose frequency manipulation to reduce the low-frequency components of the DDIM latent, which serves as the sampling startpoint, thereby enhancing original content preservation.
	\item We propose negative guidance via inversion to distance the sampling startpoint from the content of style image, thus alleviating content leakage from style image.
	\item Extensive experiments on the style transfer dataset validate that the proposed method significantly outperforms previous works both quantitatively and qualitatively.
\end{itemize}

\section{Related Work}
\label{sec:related_work}

\subsection{Diffusion-Based Text-to-Image Generation}

Recently, diffusion models have achieved significant success in image generation. Diffusion Probabilistic Models (DPMs)~\cite{DPM} are proposed to transform random noise into high-resolution images through a sequential sampling process. Many previous diffusion-based image generation works have demonstrated strong generative capabilities. Latent Diffusion Models (LDMs)~\cite{LDM1,LDM2} further revolutionize this approach by operating in a compressed latent space, using a pre-trained auto-encoder~\cite{autoEncoder1,autoEncoder2} to enhance computational efficiency and yield high-resolution images from textual descriptions. This transition to latent space not only accelerates the generation process but also improves the quality and coherence of the generated images. As text-to-image (T2I) diffusion models~\cite{BLIP-Diffusion,IP-Adapter} continue to grow in influence within the field of image generation, it has become evident that texts offer limited control over spatial and textural aspects of images. This has promoted the development of using more conditions from a reference image based on the T2I diffusion model~\cite{controlnet,IP-Adapter}. One of these particular conditions is style, which is the key focus of this paper.

\subsection{Style Transfer with T2I Models}

Style transfer is a condition-guided image generation task that applies the style of one image to another while preserving the original content. Early neural style transfer was extensively explored in deep convolutional~\cite{CNN_for_ST}, generative adversarial~\cite{CGAN,CycleGAN2017,pix2pix2017}, and transformer-based networks~\cite{Trans_for_ST_1,Trans_for_ST_2}, marking substantial progress over traditional methods based on signal processing~\cite{SP_1,SP_2}. This evolution has enabled numerous applications, particularly in advertising and marketing. With the powerful generative capacity of the T2I diffusion model, neural style transfer increasingly relies on pre-trained diffusion models to achieve style transfer. Previous methods~\cite{deadiff,InST,MOEController,stylediffusion} have relied on paired datasets with shared content but different styles to learn style concepts through reconstruction. For instance, DEADiff~\cite{deadiff} trains an additional image encoder guided by textual descriptions to separate style and content in the reference image. Although these approaches have demonstrated impressive style transfer capabilities, they are often time-consuming or fail to fully exploit the generative potential of large-scale diffusion models.

Training-free style transfer methods are gaining popularity due to their generalization and convenience. DiffStyle~\cite{DiffStyle} leverages h-space and adjusts skip connections to effectively convey style and content information, respectively. InstantStyle~\cite{InstantStyle} integrates features from a reference style image into style-specific layers, enhancing the style transfer process. However, these approaches often encounter challenges in preserving the original image layout. Methods like StyleID~\cite{StyleID} and InstantStyle plus~\cite{InstantStyle-Plus} have underscored the importance of inversion in content preservation, designing fusion operations for intermediate features between user-provided style reconstructions and other image streams. Nonetheless, these methods still face content leakage issues from style images.

To address these issues, we propose a novel, training-free method based on the sampling startpoint enhancement by frequency manipulation and negative guidance via inversion, which avoids content leakage from style images while ensuring strong content preservation.

\section{Preliminary}

\begin{figure*}[htbp]
	\centering
	\includegraphics[width=\linewidth]{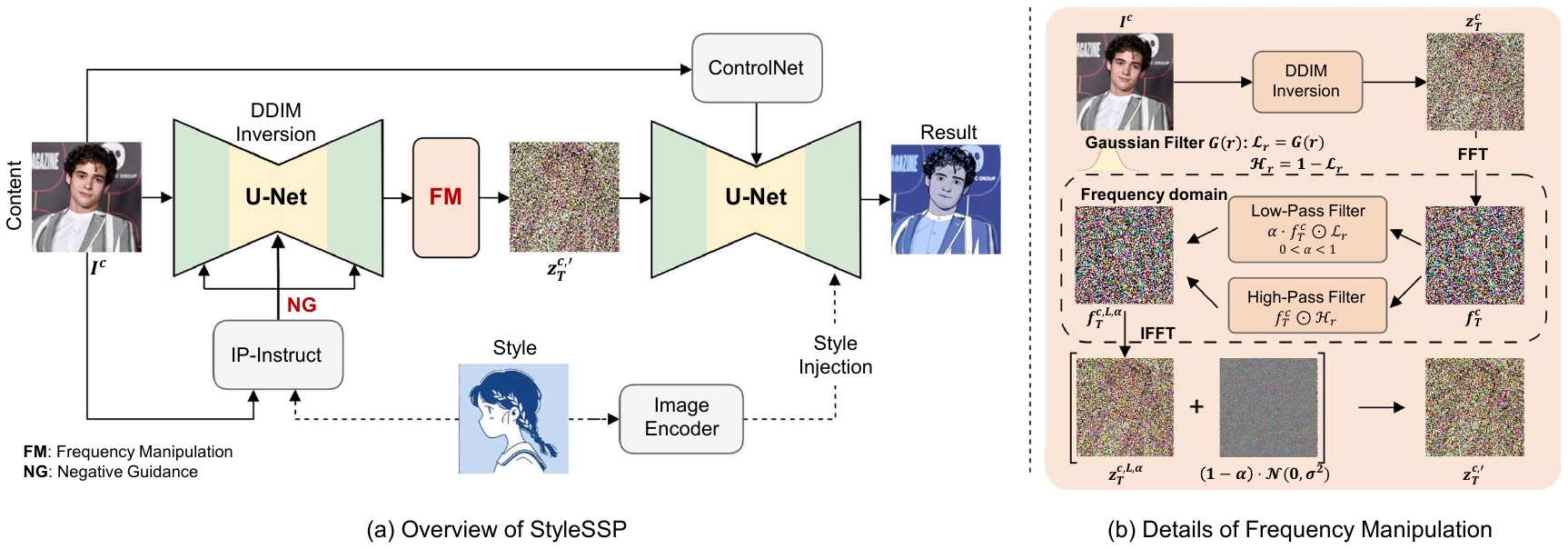}
	\caption{\textbf{Overall Framework.} (Left) Illustration of the proposed style transfer method. First, we invert the content image $I^c$ into the latent noise space as $z_T^c$. During this process, we use negative guidance (Sec.~\ref{sec: NG}) to ensure that $z_T^c$ diverges from the content information of the style image. We then apply frequency manipulation (Sec.~\ref{sec: FM}) to $z_T^c$, obtaining a low-frequency reduced latent $z_T^{c,\,'}$ as the startpoint for the sampling stage. During sampling, we follow InstantStyle's approach by injecting style features exclusively into the style-specific block and utilizing the ControlNet model to further preserve original content. (Right) Detailed explanation of frequency manipulation. We reduce the low-frequency components by a factor $\alpha$, while adding Gaussian noise proportional to $1 - \alpha$.}
	\label{fig:our_method}
\end{figure*}

\subsection{Diffusion Model}

Stable Diffusion (SD)~\cite{LDM2} is a type of latent diffusion model designed to map a random noise vector $z_t$ and a text prompt $\mathcal{P}$ to an output image $I_0$, aligning with the given conditioning prompt via cross-attention. The objective of this process is defined as:
\begin{equation}
	L = \mathbb{E}_{z_0,\epsilon \sim N(0,I),t \sim Uniform(1,T)} \| \epsilon - \epsilon_\theta(z_t, t, \mathcal{C}) \|_2^2,
	\label{Diff_Loss}
\end{equation}
where $\mathcal{C} = \varphi(\mathcal{P})$ is the embedding of text prompt generated by the text encoder $\varphi$, $t$ is the number of time steps which uniformly sampled from $\{1,...,T\}$. $\epsilon$ and $\epsilon_\theta$ represent the actual and predicted noise, respectively. The noise is gradually removed by sequentially predicting it using pre-trained diffusion model.

\noindent\textbf{Classifier-Free Guidance} (CFG)~\cite{CFG} enhances image generation quality by using a null-text embedding $\varnothing$, which corresponds to the embedding of a null text `` ", as a reference for unconditional predictions during sampling. The modified noise prediction is expressed as:
\begin{equation}
	\begin{aligned}
	\tilde{\epsilon}_\theta(z_t, t, \mathcal{C}, \varnothing)&=\epsilon_\theta(z_t, t, \varnothing)+\\
	&\omega \left( \epsilon_\theta(z_t, t, \mathcal C)-\epsilon_\theta(z_t, t, \varnothing) \right),
	\label{CFG}
	\end{aligned}
\end{equation}
where the guidance scale $\omega \geq 0$ adjusts the strength of the conditional prediction $\epsilon_\theta(z_t, t, \mathcal{C})$ against to the unconditional prediction $\epsilon_\theta(z_t, t, \varnothing)$.

\noindent\textbf{DDIM Inversion}: The Denoising Diffusion Implicit Model (DDIM)~\cite{DDIM} is a generative model that improves image synthesis efficiency and quality through a non-Markovian diffusion process, reducing the number of steps needed to generate samples. Within SD model, deterministic DDIM sampling uses a denoiser network $\epsilon_\theta$, described by:
\begin{equation}  
	z_{t-1} = \sqrt{\frac{\alpha_{t-1}}{\alpha_t}} z_t + \left( \sqrt{\frac{1}{\alpha_{t-1}} - 1} - \sqrt{\frac{1}{\alpha_t} - 1} \right) \cdot \epsilon_\theta(z_t, t),
	\label{DDIM_Sample}  
\end{equation}
where $\alpha = (\alpha_1, ..., \alpha_T) \in \mathbb{R}_{\geq 0}^T$ are hyper-parameters defining noise scales at $T$ diffusion steps. In this work, we use the publicly available SD model~\cite{SDXL}, where the diffusion forward process is applied to a latent image encoding $z_0 = E(I_0)$, and an image decoder is employed at the end of the diffusion backward process $I_0 = D(z_0)$.

By representing the DDIM sampling equation as an ordinary differential equation (ODE), the forward process can be expressed in terms of $\epsilon_\theta(z_t, t)$ by inverting the reverse diffusion process (DDIM Inversion) as follows:
\begin{equation}  
	\begin{aligned}  
		z_{t+1}^*&=\sqrt{\frac{\alpha_{t+1}}{\alpha_t}}z_t^*+\\
		&\sqrt{\alpha_{t+1}} \left( \sqrt{\frac{1}{\alpha_{t+1}}-1} - \sqrt{\frac{1}{\alpha_t}-1} \right) \cdot \epsilon_\theta(z_t^*, t).
	\end{aligned}  
	\label{DDIM Inversion}  
\end{equation}

In Eq.~\ref{DDIM Inversion}, $z_t^*$ denotes latent features during the DDIM Inversion process. Therefore, we obtain the DDIM Inversion trajectory, denoted as $[z_t^*]_{t=0}^T$. Recent works~\cite{InST,StyleID} have shown that initiating DDIM sampling from $z_T = z_T^*$ benefits to original content preservation. These findings highlight the importance of a proper startpoint for the sampling stage (denoted as $z_T$), motivating our approach to guide the inversion stage and manipulate the DDIM latent $z_T$.

\subsection{Frequency Analysis}
\label{sec: Frequency Analysis}
Inspired by FlexiEdit~\cite{flexiedit}, which highlights that high-frequency components play a more significant role in forming the object's layout than low-frequency components, we conduct a frequency analysis on the DDIM latent $z_T$ to explore frequency-domain operations that benefit to preserve the original content in style transfer. Our method separates the DDIM latent $z_T$ into high- and low-frequency components in the frequency domain as follows:
\begin{align}
	f_T^{L,\alpha}&=\alpha*f_T \odot \mathcal{L}_r+f_T\odot\mathcal{H}_r,where~\alpha\in[0,1],\\
	f_T^{H,\alpha}&=f_T \odot \mathcal{L}_r+\alpha*f_T\odot\mathcal{H}_r,where~\alpha\in[0,1],\\
	z_T^{L,\alpha}&=IFFT(f_T^{L,\alpha}),~~z_T^{H,\alpha}=IFFT(f_T^{H,\alpha}),
	\label{eq: Frequency Analysis}
\end{align}
where $FFT(\cdot)$ and $IFFT(\cdot)$ denote the 2D Fast Fourier Transform and its inverse, respectively; $f_T$ represents the frequency spectrum of $z_T$; $\mathcal{L}_r$ is a low-pass filter (e.g., Gaussian, Butterworth, or Chebyshev), and $\mathcal{H}_r = 1 - \mathcal{L}_r$ is the corresponding high-pass filter. Here, $\odot$ denotes element-wise multiplication.

\begin{figure}
	\centering
	\includegraphics[width=\linewidth]{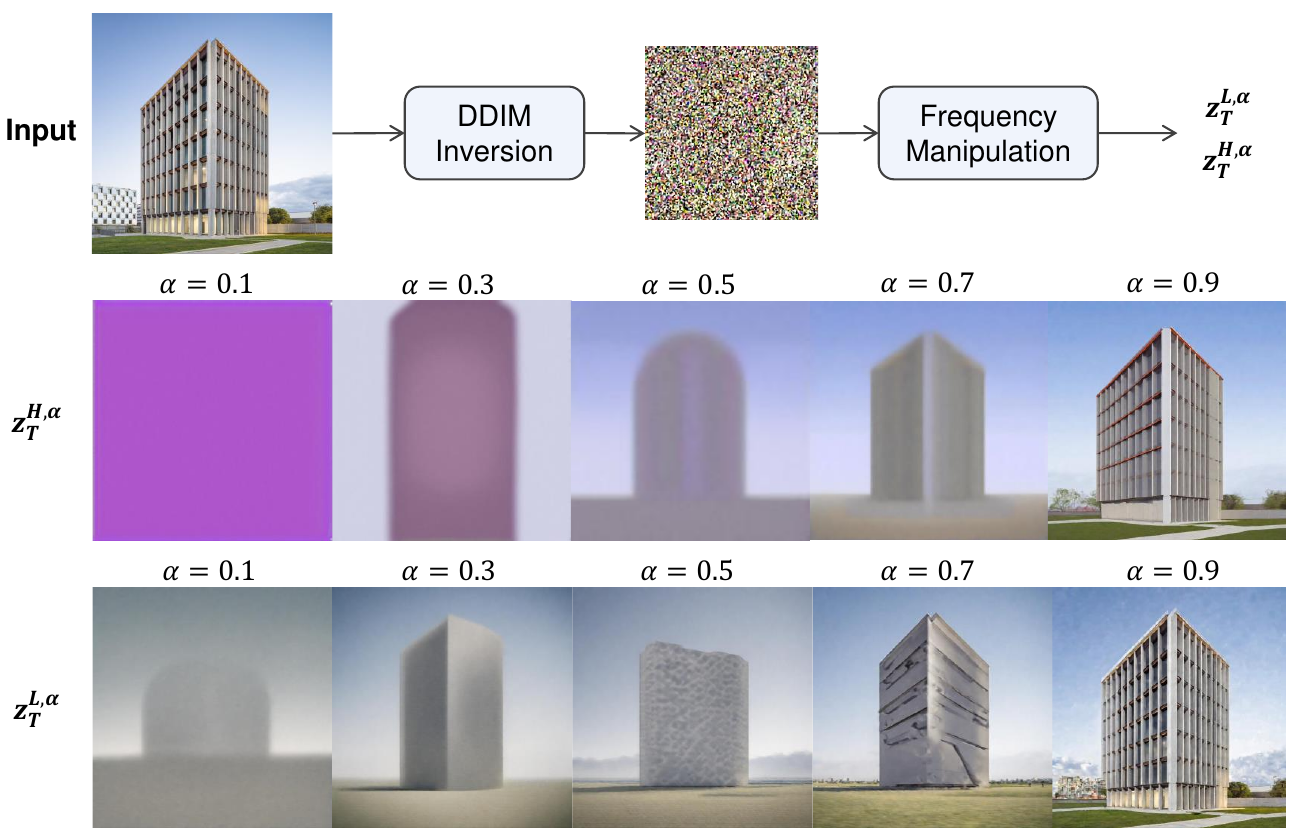}
	\caption{Reconstruction results with varying $\alpha$ values, demonstrating that high-frequency components play a critical role in the image layout, while low-frequency components contribute less to layout preservation.}
	\label{fig:FA}
\end{figure}

Since $\alpha \in [0,1]$, $z_T^{L,\alpha}$ and $z_T^{H,\alpha}$ represent low- and high-frequency reduced latents, respectively, with reduction degrees adjusted by the scale $\alpha$. In Fig.~\ref{fig:FA}, we observe that as $\alpha$ increases in reconstructions from $z_T^{H,\alpha}$, content preservation effects improve significantly. Conversely, reconstructions from $z_T^{L,\alpha}$ consistently maintain layout accuracy across varying $\alpha$ values, indicating that high-frequency components in $z_T$ are more crucial in determining the layout of image.

\section{Method}
Based on our discoveries (shown in supplementary materials Sec.~\ref{sec: startpoint effects}), which highlight the importance of a better sampling startpoint, we propose a sampling startpoint enhancement method called \textbf{StyleSSP} for training-free diffusion-based style transfer, shown in Fig.~\ref{fig:our_method}. Focusing on the problems of original content changes and content leakage from style images in current training-free methods, StyleSSP proposes two main components: (1) Frequency Manipulation (Sec.~\ref{sec: FM}) and (2) Negative Prompt Guidance via Inversion (Sec.~\ref{sec: NG}).

Let $I^c$ be a given content image whose text prompt $\mathcal{P}$ is generated by BLIP~\cite{BLIP}. Our goal is to modify the style of $I^c$ to that of style image $I^s$. The generated styled image $I^{cs}$ will maintain the content of $I^c$ while its style is consistent with $I^s$. In the following sections, we refer to the content, style, and stylized images as their encoded counterparts $z_0^c$, $z_0^s$, and $z_0^{cs}$, respectively.

\subsection{Frequency Manipulation}
\label{sec: FM}
Frequency analysis in Sec.~\ref{sec: Frequency Analysis} indicates that high-frequency components within DDIM latent $z_T^c$ of content image are more crucial in determining the layout of original image than low-frequency components. Based on this, we manipulate the frequency components of DDIM latent $z_T^{c}$ by a high-pass filter, which can achieve better preservation of original layout, resulting in improvement of details representation in the generated image.

\begin{figure}[htbp]  
	\centering  
	\includegraphics[width=\linewidth,height=2in,keepaspectratio]{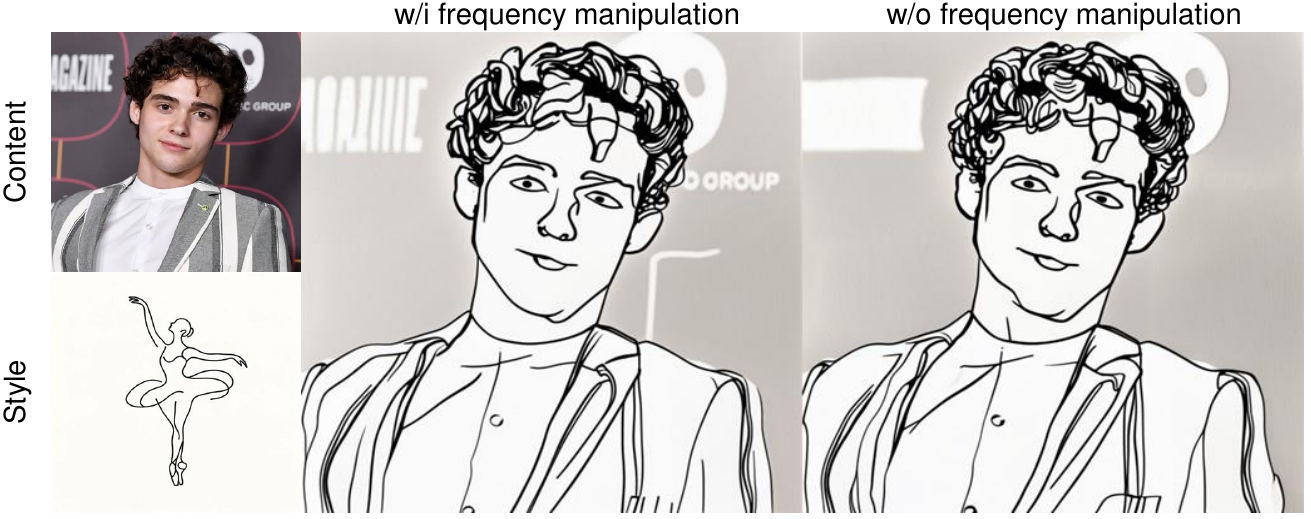}   
	\caption{Style transfer results wi/o frequency manipulation, representing the detail preservation enhancement of frequency manipulation. Result with frequency manipulation outperforms in keeping the text and lines in the background.}
	\label{fig:FM}  
\end{figure}

To this end, we first obtain the latent of content image with DDIM Inversion, and then filter the DDIM latent $z_T^c$ to get the low-frequency reduced latent $z_T^{c,L,\alpha}$, which more tightly bound with the layout of image. 

\begin{align}
	z_T^c &= \text{DDIM-Inv}(z_0^c),\\
	z_T^{c,'} &= z_T^{c, L, \alpha} + \mathcal{N}(0, \sigma^2) * (1 - \alpha),
	\label{Latent Manipulate}
\end{align}

\noindent where the definition of $z_T^{c,L,\alpha}$ is given in Eq.~\ref{eq: Frequency Analysis}, denoting the low-frequency reduced DDIM latent of content image. This procedure selectively reduces the low-frequency components by factor $\alpha$ and introduces Gaussian noise scaled by $1-\alpha$, resulting in a manipulated latent $z_T^{c,’}$. As shown in Fig.~\ref{fig:FM}, we illustrate the importance of frequency manipulation for preserving the background details of image. 

\subsection{Negative Guidance via Inversion}
\label{sec: NG}
To distance the sampling startpoint from the content of style image, we draw from insights in previous negative guidance methods. Negative prompt guidance~\cite{oconnor2023stable}, introduced in conditional generation models such as SD, allows users to specify what to exclude from generated images. This approach has gained significant attention for its effectiveness~\cite{NPA_1, NPA_2}. Specifically, when the null-text embedding $\varnothing$ in the unconditional format is replaced with an actual prompt, it represents what to remove from the generated image, leveraging the negative sign. This can be formally expressed as:

\begin{equation}
	\begin{aligned}
		\hat{\epsilon}_\theta(z_t,t,\mathcal{C_+},\mathcal{C_-})&=\epsilon_\theta(z_t,t,\mathcal{C_-})+\\
		&\omega_i(\epsilon_\theta(z_t,t,\mathcal{C_+})-\epsilon_\theta(z_t,t,\mathcal{C_-})),
	\end{aligned}
\end{equation}
where $\mathcal{C_+} = \varphi(\mathcal{P_+})$ and $\mathcal{C_-} = \varphi(\mathcal{P_-})$ are the embedding of positive text prompt $\mathcal{P_+}$ and negative text prompt $\mathcal{P_-}$, respectively. $\omega_i$ is the negative guidance scale. More details on the principles of negative prompt guidance can be found in supplementary materials Sec.~\ref{sec: principle of Negative Guidance}.

Although negative prompts provide additional control, they may interfere with the original prompt or even be disregarded~\cite{UNPG}, requiring careful tuning by users. Furthermore, the expressive capacity of text is inherently constrained, particularly for style transfer, where it is nearly impossible to comprehensively capture an image's content or precisely describe its style with words alone. These limitations substantially reduce the effectiveness of negative prompt guidance. To address this issue, we leverage the pre-trained IP-Instruct model~\cite{IP-Instruct} as a content and style extractor. The embeddings from this extractor serve as negative guidance, allowing us to overcome the challenges of accurately representing style and content information.

\begin{equation}
	\begin{aligned}
		\hat{\epsilon}_\theta(z_t,t,\mathcal{C_+},\mathcal{E}_-)&=~\epsilon_\theta(z_t,t,\mathcal{E}_-)+\\
		&\omega_i(\epsilon_\theta(z_t,t,\mathcal{C_+})-\epsilon_\theta(z_t,t,\mathcal{E}_-)),
	\end{aligned}
	\label{NG_embedding}
\end{equation}
where $\mathcal{E_-} = \text{concat}\left(\Phi(I^c)^s,\Phi(I^s)^c\right)$. $\Phi(I^s)^c$ denotes the content embedding of style image $I^s$, and $\Phi(I^c)^s$ denotes style embedding of content image $I^c$. $\Phi$ is the IP-Instruct model to extract style and content information.

\begin{figure}
	\centering
	\includegraphics[width=\linewidth]{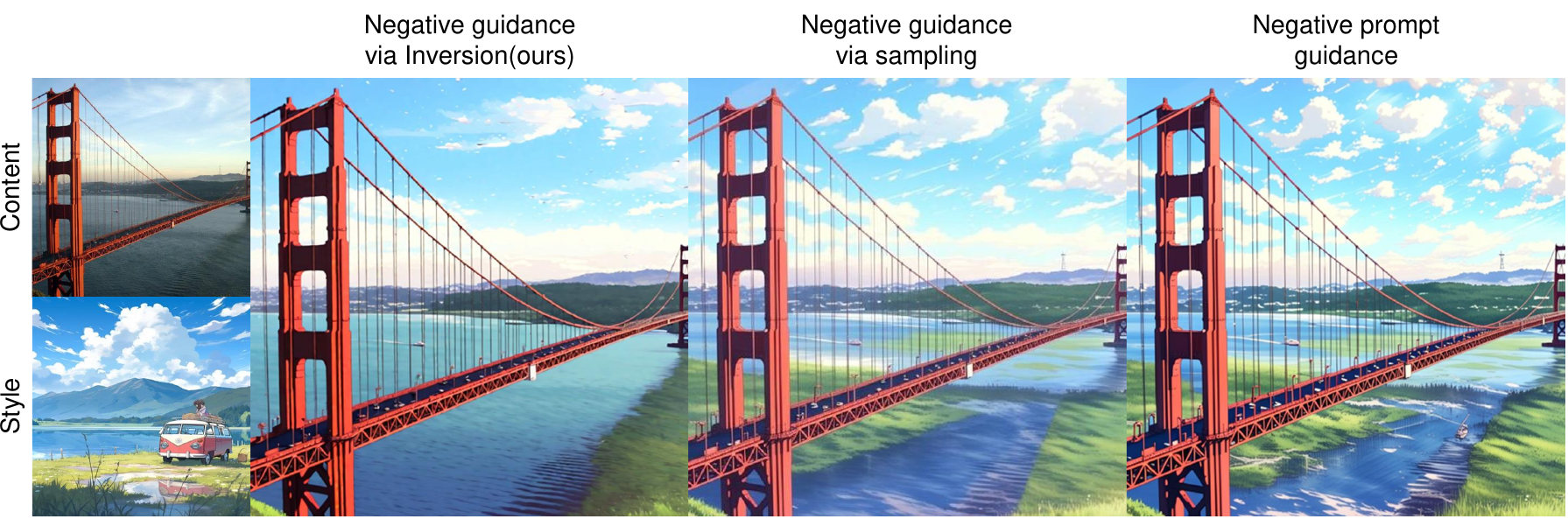}
	\caption{Illustrations of negative guidance via inversion, negative guidance in sampling step and negative prompt guidance results for style transfer. The latter two all face severe content leakage problems (the out-of-place grass on the river), while our method prevents this phenomenon. }
	\label{fig:NG}
\end{figure}

Notably, based on our significant discovery, which highlights the importance of sampling startpoint for style transfer, we innovatively employ negative guidance during DDIM Inversion. We utilize $\hat{\epsilon}_\theta(z_t,t,\mathcal{C_+},\mathcal{E}_-)$ in Eq.~\ref{NG_embedding} to replace the $\epsilon_\theta(z_t^*, t)$ in Eq.~\ref{DDIM Inversion}, presenting the predicted noises that are added into the content image gradually. As shown in Fig.~\ref{fig:NG}, the negative guidance via inversion outperforms both the traditional negative prompt guidance and the negative guidance in the sampling stage. This demonstrates that negative guidance via inversion can prevent content leakage by keeping the startpoint away from the content of style image. 

\begin{figure*}[htbp]
	\centering
	\includegraphics[width=\linewidth]{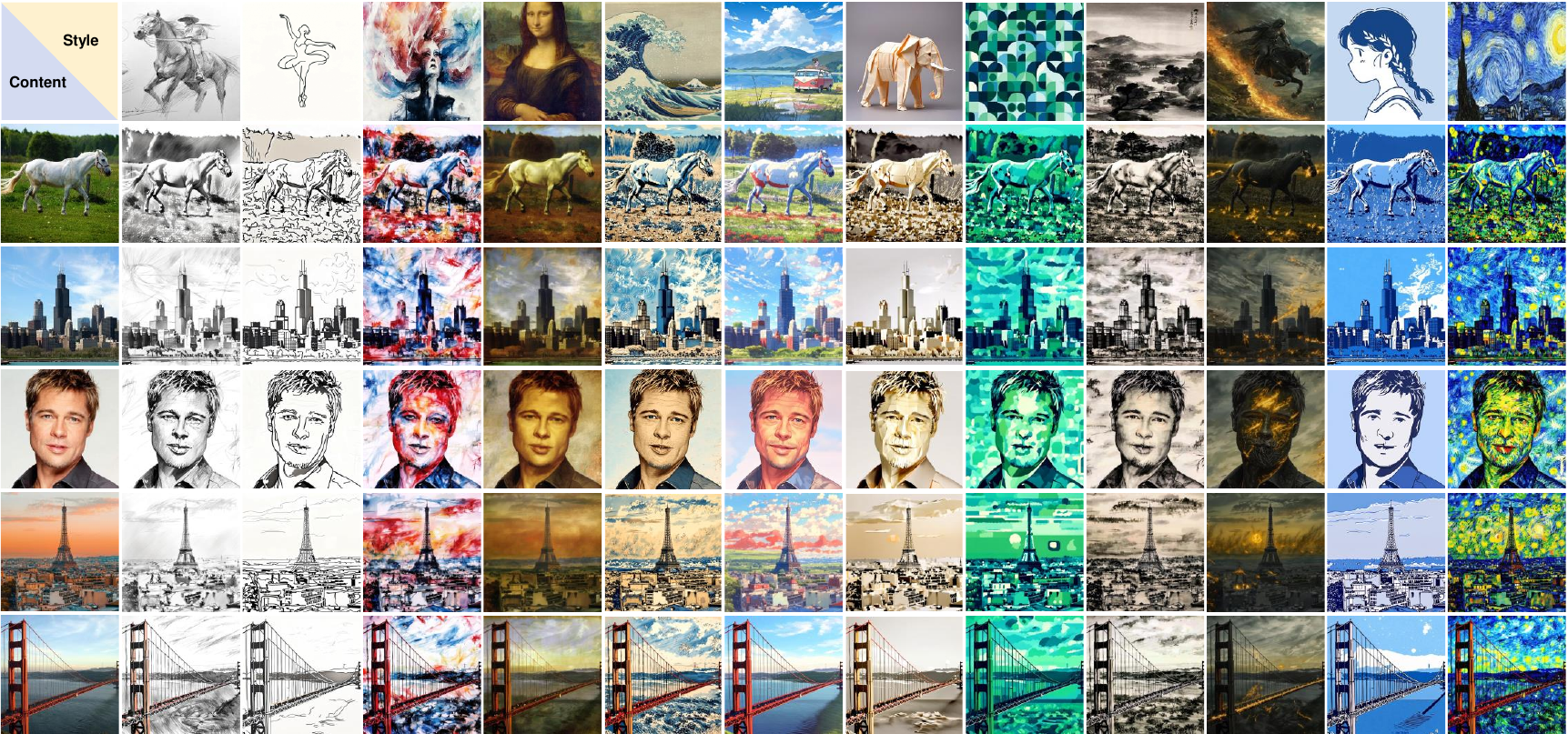}
	
	\caption{Style transfer results of style and content image pairs. Zoom in for better visualization.}
	\label{fig:our result}
\end{figure*}

\subsection{Injection \& Controlling}

\textbf{Style Injection}: Previous studies~\cite{effectoflayer_1,effectoflayer_2} have demonstrated that each layer of a deep network captures different types of semantic information, which informs the style injection strategy. This approach focuses on injecting style solely into the blocks responsible for style generation in the U-Net architecture, thereby preventing content leakage. This strategy is supported by findings from InstantStyle~\cite{InstantStyle}, which show that the first upsampling block of U-Net primarily captures style-related features such as color, material, and atmosphere. Consequently, in this work, we concentrate on injecting style features into a specific block to achieve seamless style transfer, in line with the approach used in InstantStyle.

\noindent \textbf{ControlNet for Content Preservation}: ControlNet has become one of the most widely adopted techniques for spatial conditioning, including for canny edges, depth maps, human poses, and more. In this work, we utilize ControlNet model to help preserve the layout of the content image, thereby enabling more precise control over the original content during style transfer.

\section{Experiments}

\subsection{Experimental Settings}

\begin{figure*}[htbp]
	\centering
	\includegraphics[width=\linewidth]{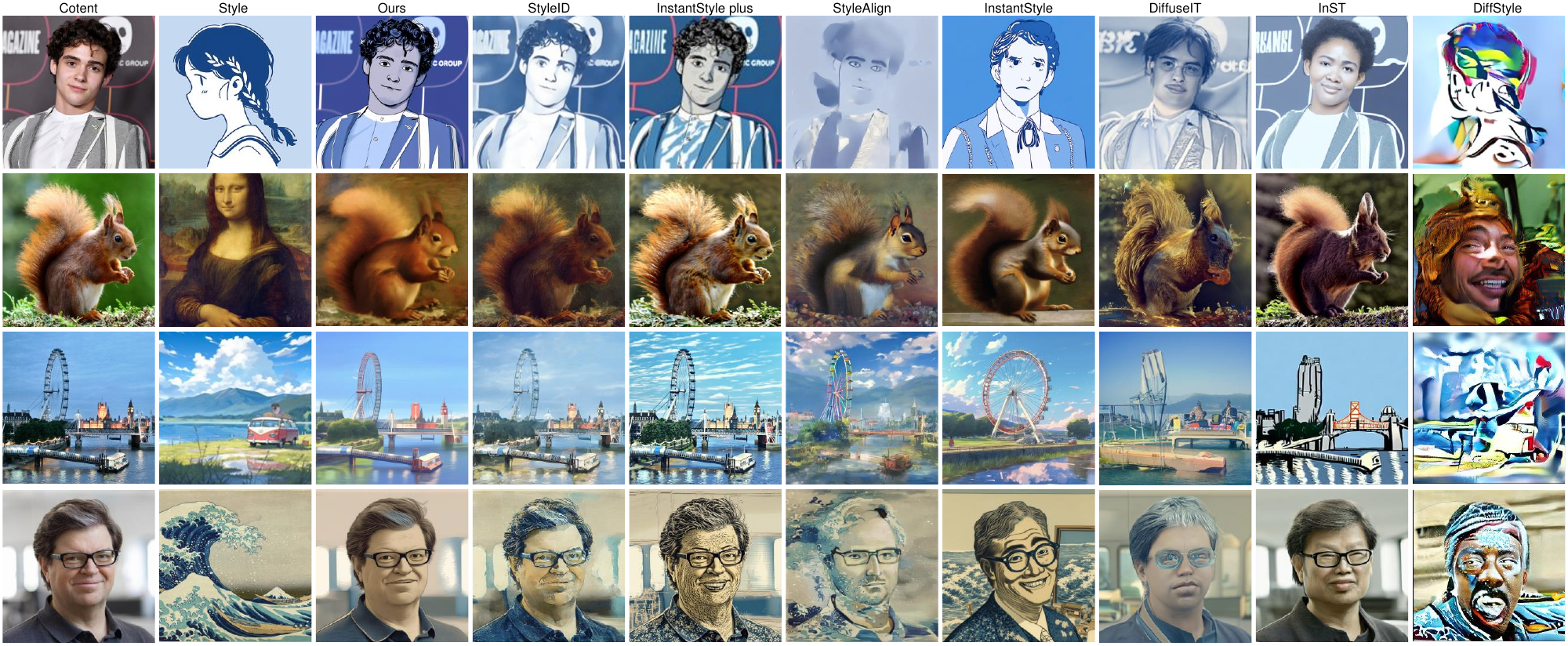}
	\caption{Qualitative comparison with previous work.}
	\label{fig:comparison}
\end{figure*}

We conduct all experiments in pre-trained Stable Diffusion XL~\cite{SDXL} and tile ControlNet~\cite{controlnet}, as well as adopt DDIM inversion and sampling with a total 50 timesteps ($t=\{1,...,50\}$). The negative guidance operates with guidance scale $\omega_i$ equal to $1.5$ while the CFG scale for sampling stage is set to $5.0$. We use the Gaussian filter with variance $\sigma$ equal to $0.3$ in frequency manipulation, and determine the scale value $\alpha$ to be $0.7$. We utilize ViT-L/14 from CLIP~\cite{CLIP} as the image encoder. All the experiments are conducted on an NVIDIA A100 GPU.

\noindent\textbf{Dataset}: Our evaluations employ content images from MS-COCO~\cite{coco} dataset and style image from WikiArt~\cite{WikiArt} dataset. For quantitative comparison, we randomly selected content and style images from each dataset, generating 800 stylized images.

\noindent\textbf{Evaluation metric}: We employ the evaluation metric ArtFID~\cite{ArtFID}, LPIPS~\cite{LIPIS} and FID~\cite{FID}, consistent with StyleID. ArtFID evaluates overall style transfer performances with consideration of both content and style preservation and also is known as strongly coinciding with human judgment, which is computed as $ArtFID = (1 + LPIPS) \cdot (1+FID)$. LPIPS measures content fidelity between the stylized image and the corresponding content image, and FID assesses the style fidelity between the stylized image and the corresponding style image.

\subsection{Qualitative Results}
Fig.~\ref{fig:our result} presents the superior style transfer results of StyleSSP across various subjects, demonstrating its robustness and versatility in adapting to diverse content and styles. The results show that our method not only performs straightforward color transfer but also captures more distinctive features, such as brush strokes and textures from the style image, leading to visually appealing style transfer effects. Additional results can be found in the supplementary materials Sec.~\ref{sec:More example}.

\subsection{Comparison with State-of-the-Art Methods}

We evaluate our proposed method by comparing it with previous state-of-the-art methods, including training-free diffusion-based methods such as StyleID~\cite{StyleID}, StyleAlign~\cite{stylediffusion}, InstantStyle plus~\cite{InstantStyle-Plus}, InstantStyle~\cite{InstantStyle}, DiffuseIT~\cite{DiffuseIT}, and DiffStyle~\cite{DiffStyle}. Additionally, we also include the optimization-based method InST~\cite{InST} in our comparison, based on the experimental settings of StyleID. 

\noindent\textbf{Quantitative Comparisons}: As shown in Tab.~\ref{tab:comparison}, our method outperforms previous style transfer methods in terms of ArtFID, FID, and LPIPS, indicating superior style resemblance and content fidelity. Several key observations can be made from this comparison. First, when compared to content preservation methods such as InstantStyle plus, StyleID, and InST, our approach achieves the best LPIPS score, demonstrating a significant improvement in content preservation. Second, our method also achieves the lowest FID, highlighting its superior style transfer performance. In summary, StyleSSP strikes an optimal balance between high-quality style transfer and precise content preservation.

\begin{table*}[htbp]
	\centering
	\begin{tabular}{@{}c|cccccccc@{}}
		\toprule
		Metric & Ours & StyleID & InstantStyle plus & StyleAlign & InstantStyle & DiffuseIT & InST & DiffStyle\\
		\midrule
		ArtFID$\downarrow$ & \textbf{21.499} & 28.801& 25.886 &36.269&37.524&40.721&40.633&41.464\\
		FID$\downarrow$ & \textbf{13.448} & 18.131 & 16.097&20.338&21.817&23.065&21.571&20.903\\
		LPIPS$\downarrow$ & \textbf{0.4881} &0.5055& 0.5140 & 0.6997&0.6446&0.6921&0.8002&0.8931\\
		\bottomrule
	\end{tabular}
	\caption{Quantitative comparison with diffusion model baselines}
	\label{tab:comparison}
\end{table*}
\noindent \textbf{Qualitative Comparisons}: Fig.~\ref{fig:comparison} presents a visual comparison between our method and previous works. Overall, our approach achieves the best visual balance between enhancing stylistic effects and preserving the original content, while effectively preventing the content leakage from style image. Several key observations can be made from this figure. First, methods without inversion exhibit significant limitations in content preservation, particularly in the background details, as shown in the $1^{\text{st}}$ row. Second, although inversion-based methods such as StyleID, InST, and InstantStyle plus present some content preservation ability, they fail to fully decouple style and content information. This results in visible content leakage in some synthesized images, especially in the $4^{\text{th}}$ row of Fig.~\ref{fig:comparison}. If users interpret the waves in the $4^{\text{th}}$ row as part of the style, we show in Sec.~\ref{sec:Additional Analysis} that content leakage can be controlled by adjusting the negative guidance scale $\omega_i$, allowing users to customize the result according to their preferences. Additional results are provided in the supplementary materials Sec.~\ref{sec:More example}.

\subsection{Ablation Study}

\begin{table}[htbp]
	\centering
	\begin{tabular}{@{}c|ccc@{}}
		\toprule
		Configuration & ArtFID$\downarrow$ & FID$\downarrow$ & LPIPS$\downarrow$\\
		\midrule
		Baseline & 26.683 & 16.205 &  0.5509\\
		+ FM & 24.112 & 15.103 & 0.4973\\
		+ NG & 26.542 & 16.128 & 0.5496\\
		StyleSSP & \textbf{21.499} & \textbf{13.448} & \textbf{0.4881}\\
		\bottomrule
	\end{tabular}
	\caption{Quantitative results from gradually increasing components with StyleSSP.}
	\label{tab:ablation}
\end{table}

\begin{figure}[htbp]
	\centering
	\includegraphics[width=\linewidth]{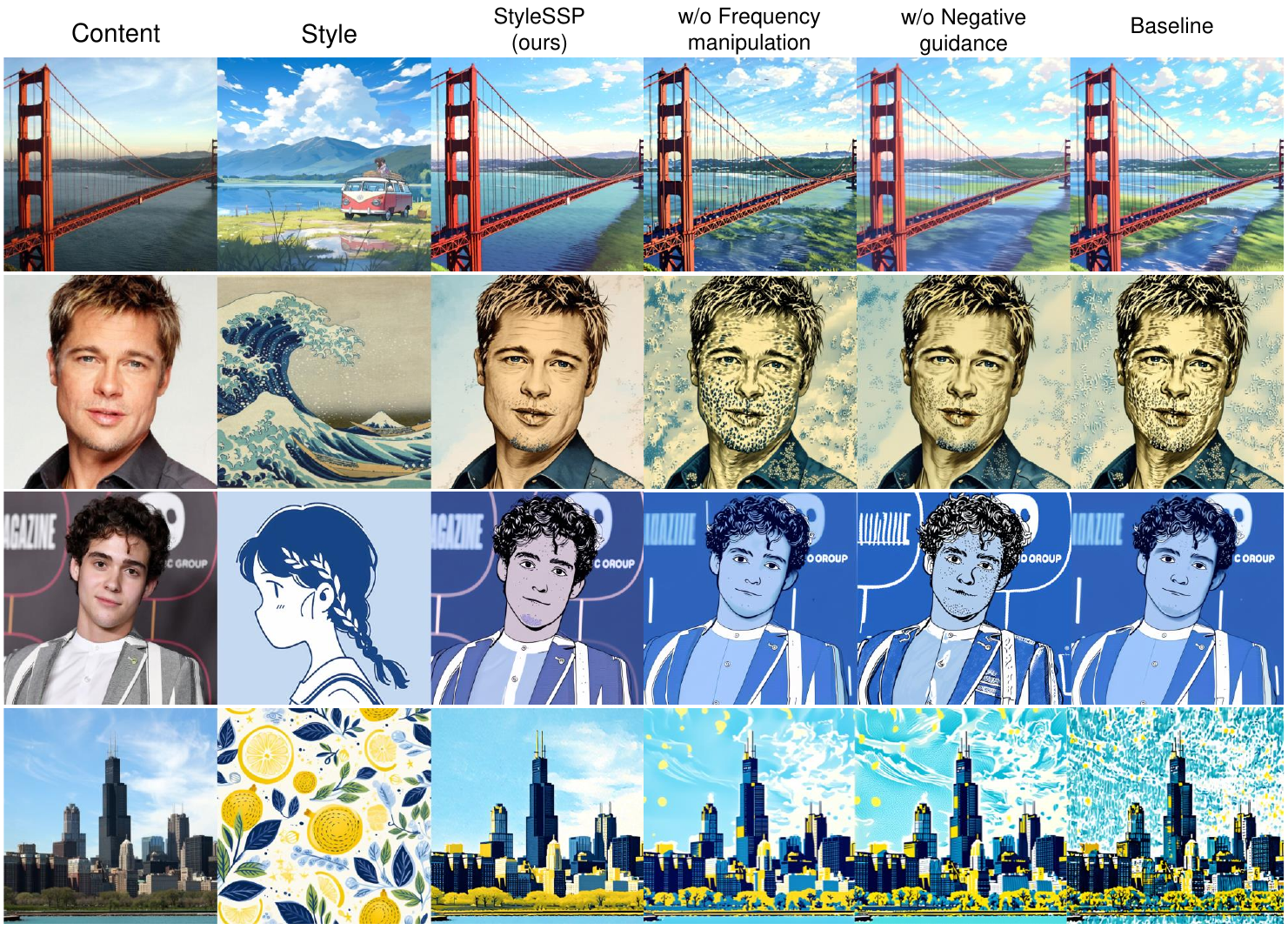}
	\caption{Qualitative comparison with ablation studies.}
	\vspace{-0.5cm}
	\label{fig:ablation study}
\end{figure}
To validate the effectiveness of the proposed components, we conduct ablation studies from both quantitative and qualitative perspectives. The baseline refers to the method without frequency manipulation (FM) and negative guidance via inversion (NG). Qualitative results, as shown in Fig.~\ref{fig:ablation study}, illustrate the effects of frequency manipulation for content preservation and negative guidance for preventing content leakage. First, referring to the $3^{\text{rd}}$ row of Fig.~\ref{fig:ablation study}, frequency manipulation significantly improves the preservation of background details in the content image. Second, referring to the $1^{\text{st}}, 2^{\text{nd}}, 4^{\text{th}}$ rows, negative guidance effectively prevents the contamination of content by style images in the generated images. By guiding the startpoint distance from the content of style image, negative guidance successfully prevents the contamination of river, human faces, and sky in the original images by the grassland, waves, and yellow dots from style images. Quantitative results shown in Tab.~\ref{tab:ablation} further demonstrate the superior performance of our proposed components. In summary, our method excels in both visual effects and quantitative metrics.

\subsection{Additional Analysis}
\label{sec:Additional Analysis}
\begin{figure}[htbp]
	\centering
	\includegraphics[width=\linewidth]{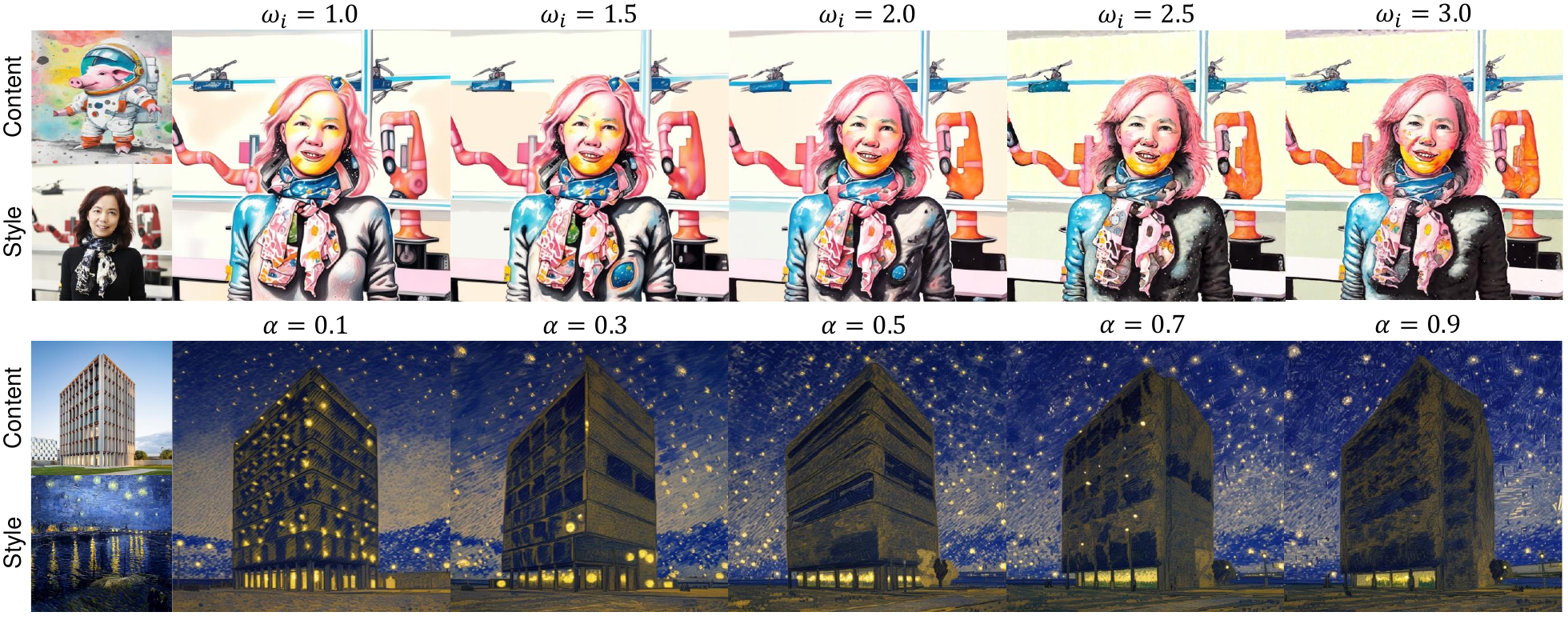}
	\caption{Visualization of the effects of negative guidance scale $\omega_i$ and frequency manipulation ratio $\alpha$.}
	\label{fig:scale}  
	\vspace{-0.2cm}
\end{figure}
We investigate the effects of different negative guidance scales $\omega_i$ and frequency manipulation ratio $\alpha$. We observe that the gradual increase of $\omega_i$ reduces the degree of content leakage from style image, as shown in Fig.~\ref{fig:scale} (top). This result further implies that negative guidance is effective in  mitigating content leakage. In addition, as shown in Fig.~\ref{fig:scale} (bottom), a lower frequency manipulation ratio $\alpha$ results in stylized images with clearer contours and more defined layouts, highlighting the importance of reducing low-frequency components in the startpoint for enhancing image structure and detail. This characteristic suggests that users can adjust the degree of contour sharpness and content leakage based on their preferences.

\section{Conclusion}
In this paper, we introduce \textbf{StyleSSP}, a novel method for sampling startpoint enhancement in training-free diffusion-based style transfer. To the best of our knowledge, we are the first to emphasize the importance of the sampling startpoint in style transfer. We identify two key challenges in training-free methods: changes of original content and content leakage from style images. These issues stem primarily from the absence of targeted training for style extraction and constraints on content layout. To address these issues, we propose two components for optimizing the sampling startpoint: (1) frequency manipulation for improved content preservation, and (2) negative guidance via inversion to prevent content leakage. Empirical results demonstrate that StyleSSP effectively mitigates original content changes and content leakage from style image while achieving superior style transfer performance. Comparison experiments show that StyleSSP outperforms previous methods both qualitatively and quantitatively. Future work could explore regionally-aware startpoint manipulation techniques to further enhance objective-level stylization.


\small
\bibliographystyle{ieeenat_fullname}
\bibliography{main}

\begin{thebibliography}{56}
\providecommand{\natexlab}[1]{#1}
\providecommand{\url}[1]{\texttt{#1}}
\expandafter\ifx\csname urlstyle\endcsname\relax
  \providecommand{\doi}[1]{doi: #1}\else
  \providecommand{\doi}{doi: \begingroup \urlstyle{rm}\Url}\fi

\bibitem[Armandpour et~al.(2023)Armandpour, Sadeghian, Zheng, Sadeghian, and
  Zhou]{NPA_1}
Mohammadreza Armandpour, Ali Sadeghian, Huangjie Zheng, Amir Sadeghian, and
  Mingyuan Zhou.
\newblock Re-imagine the negative prompt algorithm: Transform 2d diffusion into
  3d, alleviate janus problem and beyond, 2023.

\bibitem[Ban et~al.(2024)Ban, Wang, Zhou, Cheng, Gong, and Hsieh]{UNPG}
Yuanhao Ban, Ruochen Wang, Tianyi Zhou, Minhao Cheng, Boqing Gong, and Cho-Jui
  Hsieh.
\newblock Understanding the impact of negative prompts: When and how do they
  take effect?, 2024.

\bibitem[Bodur et~al.(2024)Bodur, Gundogdu, Bhattarai, Kim, Donoser, and
  Bazzani]{Edit_3}
Rumeysa Bodur, Erhan Gundogdu, Binod Bhattarai, Tae-Kyun Kim, Michael Donoser,
  and Loris Bazzani.
\newblock iedit: Localised text-guided image editing with weak supervision.
\newblock In \emph{Proceedings of the IEEE/CVF Conference on Computer Vision
  and Pattern Recognition (CVPR) Workshops}, pages 7426--7435, 2024.

\bibitem[Chung et~al.(2024)Chung, Hyun, and Heo]{StyleID}
Jiwoo Chung, Sangeek Hyun, and Jae-Pil Heo.
\newblock Style injection in diffusion: A training-free approach for adapting
  large-scale diffusion models for style transfer.
\newblock In \emph{Proceedings of the IEEE/CVF Conference on Computer Vision
  and Pattern Recognition (CVPR)}, pages 8795--8805, 2024.

\bibitem[Efros and Freeman(2001)]{SP_1}
Alexei~A. Efros and William~T. Freeman.
\newblock Image quilting for texture synthesis and transfer.
\newblock In \emph{Proceedings of the 28th Annual Conference on Computer
  Graphics and Interactive Techniques}, page 341–346, New York, NY, USA,
  2001. Association for Computing Machinery.

\bibitem[Gatys et~al.(2015)Gatys, Ecker, and Bethge]{CNN_for_ST}
Leon~A. Gatys, Alexander~S. Ecker, and Matthias Bethge.
\newblock Texture synthesis using convolutional neural networks, 2015.

\bibitem[Geng et~al.(2023)Geng, Yang, Hang, Li, Gu, Zhang, Bao, Zhang, Hu,
  Chen, and Guo]{Edit_2}
Zigang Geng, Binxin Yang, Tiankai Hang, Chen Li, Shuyang Gu, Ting Zhang,
  Jianmin Bao, Zheng Zhang, Han Hu, Dong Chen, and Baining Guo.
\newblock Instructdiffusion: {A} generalist modeling interface for vision
  tasks.
\newblock \emph{CoRR}, abs/2309.03895, 2023.

\bibitem[Geng et~al.(2024)Geng, Yang, Hang, Li, Gu, Zhang, Bao, Zhang, Li, Hu,
  Chen, and Guo]{Edit_4}
Zigang Geng, Binxin Yang, Tiankai Hang, Chen Li, Shuyang Gu, Ting Zhang,
  Jianmin Bao, Zheng Zhang, Houqiang Li, Han Hu, Dong Chen, and Baining Guo.
\newblock Instructdiffusion: A generalist modeling interface for vision tasks.
\newblock In \emph{2024 IEEE/CVF Conference on Computer Vision and Pattern
  Recognition (CVPR)}, pages 12709--12720, 2024.

\bibitem[Han et~al.(2024)Han, Wang, Zhang, Hu, Cheng, Fu, and Zhang]{EMMA}
Yucheng Han, Rui Wang, Chi Zhang, Juntao Hu, Pei Cheng, Bin Fu, and Hanwang
  Zhang.
\newblock Emma: Your text-to-image diffusion model can secretly accept
  multi-modal prompts, 2024.

\bibitem[Ho(2022)]{CFG}
Jonathan Ho.
\newblock Classifier-free diffusion guidance.
\newblock \emph{ArXiv}, abs/2207.12598, 2022.

\bibitem[Isola et~al.(2017)Isola, Zhu, Zhou, and Efros]{pix2pix2017}
Phillip Isola, Jun-Yan Zhu, Tinghui Zhou, and Alexei~A Efros.
\newblock Image-to-image translation with conditional adversarial networks.
\newblock \emph{CVPR}, 2017.

\bibitem[Jeong et~al.(2024)Jeong, Kwon, and Uh]{DiffStyle}
Jaeseok Jeong, Mingi Kwon, and Youngjung Uh.
\newblock Training-free content injection using h-space in diffusion models,
  2024.

\bibitem[Ju et~al.(2024)Ju, Liu, Wang, Bian, Shan, and Xu]{brushnet}
Xuan Ju, Xian Liu, Xintao Wang, Yuxuan Bian, Ying Shan, and Qiang Xu.
\newblock Brushnet: A plug-and-play image inpainting model with decomposed
  dual-branch diffusion, 2024.

\bibitem[Kawar et~al.(2023)Kawar, Zada, Lang, Tov, Chang, Dekel, Mosseri, and
  Irani]{Edit_1}
Bahjat Kawar, Shiran Zada, Oran Lang, Omer Tov, Huiwen Chang, Tali Dekel, Inbar
  Mosseri, and Michal Irani.
\newblock Imagic: Text-based real image editing with diffusion models.
\newblock In \emph{Conference on Computer Vision and Pattern Recognition 2023},
  2023.

\bibitem[Kolkin et~al.(2019)Kolkin, Salavon, and Shakhnarovich]{Trans_for_ST_1}
Nicholas Kolkin, Jason Salavon, and Greg Shakhnarovich.
\newblock Style transfer by relaxed optimal transport and self-similarity,
  2019.

\bibitem[Koo et~al.(2024)Koo, Yoon, Hong, and Yoo]{flexiedit}
Gwanhyeong Koo, Sunjae Yoon, Ji~Woo Hong, and Chang~D Yoo.
\newblock Flexiedit: Frequency-aware latent refinement for enhanced non-rigid
  editing.
\newblock \emph{arXiv preprint arXiv:2407.17850}, 2024.

\bibitem[Kwon and Ye(2023)]{DiffuseIT}
Gihyun Kwon and Jong~Chul Ye.
\newblock Diffusion-based image translation using disentangled style and
  content representation, 2023.

\bibitem[Li et~al.(2023{\natexlab{a}})Li, Li, and Hoi]{BLIP-Diffusion}
Dongxu Li, Junnan Li, and Steven C.~H. Hoi.
\newblock Blip-diffusion: Pre-trained subject representation for controllable
  text-to-image generation and editing, 2023{\natexlab{a}}.

\bibitem[Li et~al.(2022)Li, Li, Xiong, and Hoi]{BLIP}
Junnan Li, Dongxu Li, Caiming Xiong, and Steven Hoi.
\newblock Blip: Bootstrapping language-image pre-training for unified
  vision-language understanding and generation, 2022.

\bibitem[Li et~al.(2023{\natexlab{b}})Li, van~de Weijer, Hu, Khan, Hou, Wang,
  and Yang]{stylediffusion}
Senmao Li, Joost van~de Weijer, Taihang Hu, Fahad~Shahbaz Khan, Qibin Hou,
  Yaxing Wang, and Jian Yang.
\newblock Stylediffusion: Prompt-embedding inversion for text-based editing.
\newblock \emph{arXiv preprint arXiv:2303.15649}, 2023{\natexlab{b}}.

\bibitem[Li et~al.(2024)Li, Chen, and Lu]{MOEController}
Sijia Li, Chen Chen, and Haonan Lu.
\newblock Moecontroller: Instruction-based arbitrary image manipulation with
  mixture-of-expert controllers, 2024.

\bibitem[Lin et~al.(2015)Lin, Maire, Belongie, Bourdev, Girshick, Hays, Perona,
  Ramanan, Zitnick, and Dollár]{coco}
Tsung-Yi Lin, Michael Maire, Serge Belongie, Lubomir Bourdev, Ross Girshick,
  James Hays, Pietro Perona, Deva Ramanan, C.~Lawrence Zitnick, and Piotr
  Dollár.
\newblock Microsoft coco: Common objects in context, 2015.

\bibitem[Mahendran and Vedaldi(2014)]{effectoflayer_2}
Aravindh Mahendran and Andrea Vedaldi.
\newblock Understanding deep image representations by inverting them, 2014.

\bibitem[Men et~al.(2018)Men, Lian, Tang, and Xiao]{SP_2}
Yifang Men, Zhouhui Lian, Yingmin Tang, and Jianguo Xiao.
\newblock A common framework for interactive texture transfer.
\newblock In \emph{Proceedings of the IEEE Conference on Computer Vision and
  Pattern Recognition}, pages 6353--6362, 2018.

\bibitem[Mirza and Osindero(2014)]{CGAN}
Mehdi Mirza and Simon Osindero.
\newblock Conditional generative adversarial nets.
\newblock \emph{CoRR}, abs/1411.1784, 2014.

\bibitem[Mokady et~al.(2022)Mokady, Hertz, Aberman, Pritch, and Cohen-Or]{NTI}
Ron Mokady, Amir Hertz, Kfir Aberman, Yael Pritch, and Daniel Cohen-Or.
\newblock Null-text inversion for editing real images using guided diffusion
  models, 2022.

\bibitem[Nichol et~al.(2022)Nichol, Dhariwal, Ramesh, Shyam, Mishkin, McGrew,
  Sutskever, and Chen]{T2I_1}
Alex Nichol, Prafulla Dhariwal, Aditya Ramesh, Pranav Shyam, Pamela Mishkin,
  Bob McGrew, Ilya Sutskever, and Mark Chen.
\newblock Glide: Towards photorealistic image generation and editing with
  text-guided diffusion models, 2022.

\bibitem[O’Connor(2023)]{oconnor2023stable}
Ryan O’Connor.
\newblock Stable diffusion 1 vs 2: What you need to know.
\newblock
  \url{https://www.assemblyai.com/blog/stable-diffusion-1-vs-2-what-you-need-to-know},
  2023.

\bibitem[Park and Lee(2019)]{Trans_for_ST_2}
Dae~Young Park and Kwang~Hee Lee.
\newblock Arbitrary style transfer with style-attentional networks, 2019.

\bibitem[Podell et~al.(2023)Podell, English, Lacey, Blattmann, Dockhorn,
  Müller, Penna, and Rombach]{SDXL}
Dustin Podell, Zion English, Kyle Lacey, Andreas Blattmann, Tim Dockhorn, Jonas
  Müller, Joe Penna, and Robin Rombach.
\newblock Sdxl: Improving latent diffusion models for high-resolution image
  synthesis, 2023.

\bibitem[Qi et~al.(2024)Qi, Fang, Wu, Xie, Liu, Chen, He, and Zhang]{deadiff}
Tianhao Qi, Shancheng Fang, Yanze Wu, Hongtao Xie, Jiawei Liu, Lang Chen, Qian
  He, and Yongdong Zhang.
\newblock Deadiff: An efficient stylization diffusion model with disentangled
  representations.
\newblock \emph{arXiv preprint arXiv:2403.06951}, 2024.

\bibitem[Qiu et~al.(2024)Qiu, Xia, Zhang, He, Wang, Shan, and Liu]{FreeNoise}
Haonan Qiu, Menghan Xia, Yong Zhang, Yingqing He, Xintao Wang, Ying Shan, and
  Ziwei Liu.
\newblock Freenoise: Tuning-free longer video diffusion via noise rescheduling,
  2024.

\bibitem[Radford et~al.(2021)Radford, Kim, Hallacy, Ramesh, Goh, Agarwal,
  Sastry, Askell, Mishkin, Clark, Krueger, and Sutskever]{autoEncoder1}
Alec Radford, Jong~Wook Kim, Chris Hallacy, Aditya Ramesh, Gabriel Goh,
  Sandhini Agarwal, Girish Sastry, Amanda Askell, Pamela Mishkin, Jack Clark,
  Gretchen Krueger, and Ilya Sutskever.
\newblock Learning transferable visual models from natural language
  supervision, 2021.

\bibitem[Raffel et~al.(2017)Raffel, Luong, Liu, Weiss, and Eck]{autoEncoder2}
Colin Raffel, Minh-Thang Luong, Peter~J. Liu, Ron~J. Weiss, and Douglas Eck.
\newblock Online and linear-time attention by enforcing monotonic alignments,
  2017.

\bibitem[Ramesh et~al.(2022{\natexlab{a}})Ramesh, Dhariwal, Nichol, Chu, and
  Chen]{CLIP}
Aditya Ramesh, Prafulla Dhariwal, Alex Nichol, Casey Chu, and Mark Chen.
\newblock Hierarchical text-conditional image generation with clip latents,
  2022{\natexlab{a}}.

\bibitem[Ramesh et~al.(2022{\natexlab{b}})Ramesh, Dhariwal, Nichol, Chu, and
  Chen]{LDM1}
Aditya Ramesh, Prafulla Dhariwal, Alex Nichol, Casey Chu, and Mark Chen.
\newblock Hierarchical text-conditional image generation with clip latents,
  2022{\natexlab{b}}.

\bibitem[Rombach et~al.(2021)Rombach, Blattmann, Lorenz, Esser, and
  Ommer]{T2I_2}
Robin Rombach, Andreas Blattmann, Dominik Lorenz, Patrick Esser, and Björn
  Ommer.
\newblock High-resolution image synthesis with latent diffusion models, 2021.

\bibitem[Rombach et~al.(2022)Rombach, Blattmann, Lorenz, Esser, and
  Ommer]{LDM2}
Robin Rombach, Andreas Blattmann, Dominik Lorenz, Patrick Esser, and Björn
  Ommer.
\newblock High-resolution image synthesis with latent diffusion models, 2022.

\bibitem[Rowles et~al.(2024)Rowles, Vainer, Nigris, Elizarov, Kutsy, and
  Donné]{IP-Instruct}
Ciara Rowles, Shimon Vainer, Dante~De Nigris, Slava Elizarov, Konstantin Kutsy,
  and Simon Donné.
\newblock Ipadapter-instruct: Resolving ambiguity in image-based conditioning
  using instruct prompts, 2024.

\bibitem[Saharia et~al.(2022)Saharia, Chan, Saxena, Li, Whang, Denton,
  Ghasemipour, Ayan, Mahdavi, Lopes, Salimans, Ho, Fleet, and Norouzi]{T2I_3}
Chitwan Saharia, William Chan, Saurabh Saxena, Lala Li, Jay Whang, Emily
  Denton, Seyed Kamyar~Seyed Ghasemipour, Burcu~Karagol Ayan, S.~Sara Mahdavi,
  Rapha~Gontijo Lopes, Tim Salimans, Jonathan Ho, David~J Fleet, and Mohammad
  Norouzi.
\newblock Photorealistic text-to-image diffusion models with deep language
  understanding, 2022.

\bibitem[Seitzer(2020)]{FID}
Maximilian Seitzer.
\newblock {pytorch-fid: FID Score for PyTorch}.
\newblock \url{https://github.com/mseitzer/pytorch-fid}, 2020.
\newblock Version 0.3.0.

\bibitem[Sohl{-}Dickstein et~al.(2015)Sohl{-}Dickstein, Weiss, Maheswaranathan,
  and Ganguli]{DPM}
Jascha Sohl{-}Dickstein, Eric~A. Weiss, Niru Maheswaranathan, and Surya
  Ganguli.
\newblock Deep unsupervised learning using nonequilibrium thermodynamics.
\newblock \emph{CoRR}, abs/1503.03585, 2015.

\bibitem[Song et~al.(2020)Song, Meng, and Ermon]{DDIM}
Jiaming Song, Chenlin Meng, and Stefano Ermon.
\newblock Denoising diffusion implicit models.
\newblock \emph{arXiv:2010.02502}, 2020.

\bibitem[Tan et~al.(2018)Tan, Chan, Aguirre, and Tanaka]{WikiArt}
Wei~Ren Tan, Chee~Seng Chan, Hernan Aguirre, and Kiyoshi Tanaka.
\newblock Improved artgan for conditional synthesis of natural image and
  artwork, 2018.

\bibitem[Wang et~al.(2024{\natexlab{a}})Wang, Wang, Bai, Qin, and
  Chen]{InstantStyle}
Haofan Wang, Qixun Wang, Xu Bai, Zekui Qin, and Anthony Chen.
\newblock Instantstyle: Free lunch towards style-preserving in text-to-image
  generation.
\newblock \emph{arXiv preprint arXiv:2404.02733}, 2024{\natexlab{a}}.

\bibitem[Wang et~al.(2024{\natexlab{b}})Wang, Xing, Huang, Ai, Wang, and
  Bai]{InstantStyle-Plus}
Haofan Wang, Peng Xing, Renyuan Huang, Hao Ai, Qixun Wang, and Xu Bai.
\newblock Instantstyle-plus: Style transfer with content-preserving in
  text-to-image generation.
\newblock \emph{arXiv preprint arXiv:2407.00788}, 2024{\natexlab{b}}.

\bibitem[Woolf(2023)]{NPA_2}
Max Woolf.
\newblock Stable diffusion 2.0 and the importance of negative prompts for good
  results.
\newblock
  \url{https://minimaxir.com/2022/11/stable-diffusion-negative-prompt/}, 2023.

\bibitem[Wright and Ommer(2022)]{ArtFID}
Matthias Wright and Bj{\"o}rn Ommer.
\newblock Artfid: Quantitative evaluation of neural style transfer.
\newblock \emph{GCPR}, 2022.

\bibitem[Wu et~al.(2021)Wu, Nitzan, Shechtman, and Lischinski]{styleAlign}
Zongze Wu, Yotam Nitzan, Eli Shechtman, and Dani Lischinski.
\newblock Stylealign: Analysis and applications of aligned stylegan models.
\newblock \emph{arXiv preprint arXiv:2110.11323}, 2021.

\bibitem[Xu et~al.(2024)Xu, Wang, Xiao, Liu, and Chen]{FreeTuner}
Youcan Xu, Zhen Wang, Jun Xiao, Wei Liu, and Long Chen.
\newblock Freetuner: Any subject in any style with training-free diffusion,
  2024.

\bibitem[Ye et~al.(2023)Ye, Zhang, Liu, Han, and Yang]{IP-Adapter}
Hu Ye, Jun Zhang, Sibo Liu, Xiao Han, and Wei Yang.
\newblock Ip-adapter: Text compatible image prompt adapter for text-to-image
  diffusion models, 2023.

\bibitem[Yosinski et~al.(2014)Yosinski, Clune, Bengio, and
  Lipson]{effectoflayer_1}
Jason Yosinski, Jeff Clune, Yoshua Bengio, and Hod Lipson.
\newblock How transferable are features in deep neural networks?, 2014.

\bibitem[Zhang et~al.(2023{\natexlab{a}})Zhang, Rao, and Agrawala]{controlnet}
Lvmin Zhang, Anyi Rao, and Maneesh Agrawala.
\newblock Adding conditional control to text-to-image diffusion models,
  2023{\natexlab{a}}.

\bibitem[Zhang et~al.(2018)Zhang, Isola, Efros, Shechtman, and Wang]{LIPIS}
Richard Zhang, Phillip Isola, Alexei~A. Efros, Eli Shechtman, and Oliver Wang.
\newblock The unreasonable effectiveness of deep features as a perceptual
  metric, 2018.

\bibitem[Zhang et~al.(2023{\natexlab{b}})Zhang, Huang, Tang, Huang, Ma, Dong,
  and Xu]{InST}
Yuxin Zhang, Nisha Huang, Fan Tang, Haibin Huang, Chongyang Ma, Weiming Dong,
  and Changsheng Xu.
\newblock Inversion-based style transfer with diffusion models.
\newblock In \emph{Proceedings of the IEEE/CVF Conference on Computer Vision
  and Pattern Recognition (CVPR)}, pages 10146--10156, 2023{\natexlab{b}}.

\bibitem[Zhu et~al.(2017)Zhu, Park, Isola, and Efros]{CycleGAN2017}
Jun-Yan Zhu, Taesung Park, Phillip Isola, and Alexei~A Efros.
\newblock Unpaired image-to-image translation using cycle-consistent
  adversarial networkss.
\newblock In \emph{Computer Vision (ICCV), 2017 IEEE International Conference
  on}, 2017.

\end{thebibliography}

\clearpage
\setcounter{page}{1}
\maketitlesupplementary

\section{Appendix}

\subsection{Startpoint Impact Analysis}
\label{sec: startpoint effects}
\begin{figure*}
	\centering
	\includegraphics[width=\linewidth]{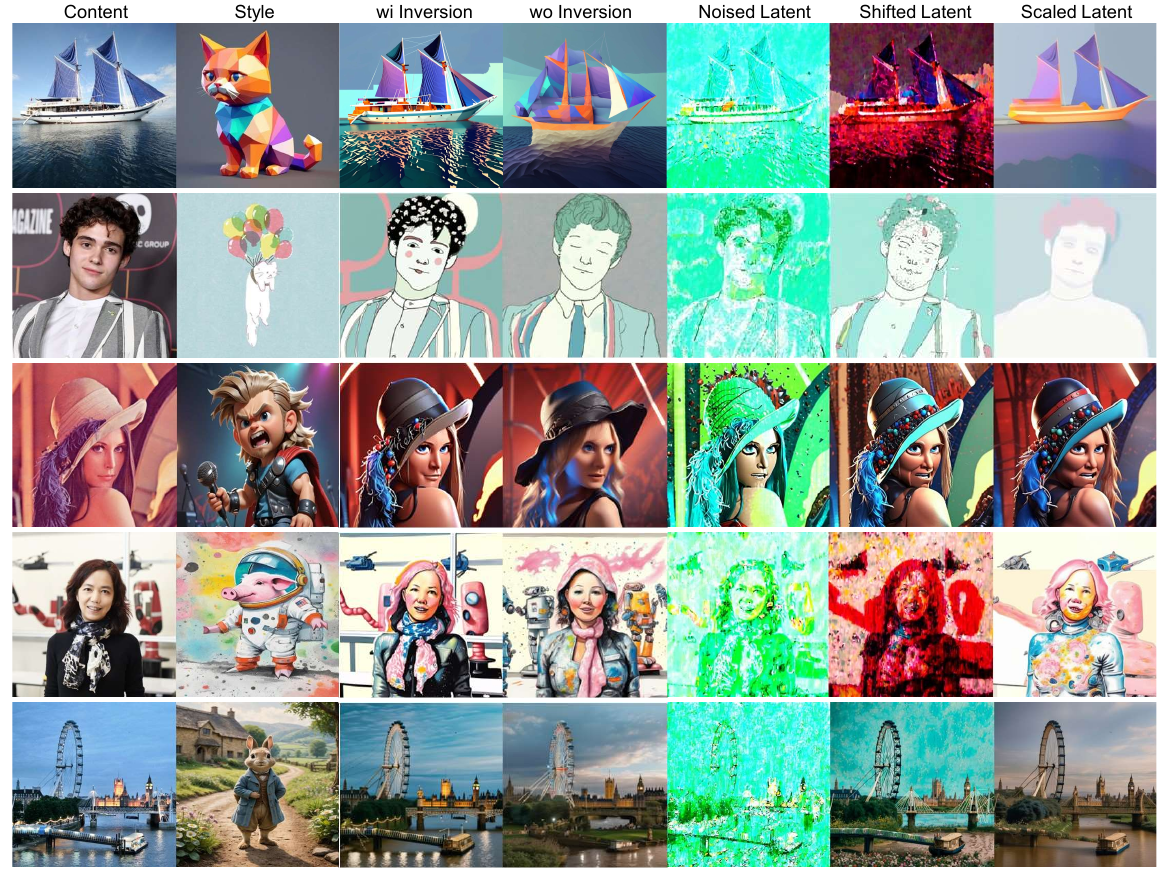}
	\caption{Illustrations of style transfer results based on various startpoints. As shown in this figure, startpoint manipulations yield significant changes in both image hue and content representation, underscoring the crucial role of the sampling startpoint in style transfer. All results are generated with ControlNet as an additional content controller.}
	\label{fig:startpoint}
\end{figure*}

Given that StyleSSP is specifically designed to enhance the sampling startpoint, we place primary emphasis on the importance of the startpoint in style transfer. We demonstrate how minor modifications to the startpoint can significantly influence style transfer results. As shown in Fig.~\ref{fig:startpoint}, we present several style transfer results. The titles in the figure — ``wi Inversion," ``wo Inversion," ``Noised Latent," ``Shifted Latent," and ``Scaled Latent" — correspond to the startpoints $z_T$, $z_r$, $z_T^n$, $z_T^{sh}$, and $z_T^{sa}$, respectively. Their formulations are as follows:
\begin{equation}
	\begin{aligned}
		z_r &\sim \mathcal{N}(0,\textbf{I})\\
		z_T^n &= z_T + \mathcal{N}(0,\textbf{I})\\
		z_T^{sh} &= z_T + \textbf{U}(-0.5,0.5)\\
		z_T^{sa} &= z_T \times \textbf{U}(0.5,1)
	\end{aligned}
\end{equation}
where $z_T$ is the DDIM latent of the content image, $\mathcal{N}$ represents a Gaussian distribution, and $U(-0.5,0.5)$ and $U(0.5,1)$ indicate uniformly random values selected within the ranges -0.5 to 0.5 and 0.5 to 1.0, respectively.


As illustrated in Fig.~\ref{fig:startpoint}, manipulations of the sampling startpoint make a significant impact on the results of style transfer, resulting in notable changes in both the image hue and the content representation. Note that the following results are all conducted with ControlNet as an additional content controller. Several key observations can be made from this figure. 

First, referring to the $3^{\text{rd}}$ and $4^{\text{th}}$ columns in this figure, using the DDIM latent $z_T$ extracted from the content image as the sampling startpoint results in remarkably better content preservation compared to using random Gaussian noise as the startpoint. This finding motivates us to adopt DDIM inversion as the first step in our method, as is done in many inversion-based methods~\cite{InstantStyle-Plus, InST, StyleID}. 

Second, we attempted minor modifications to the DDIM latent $z_T$. Referring to the $3^{\text{rd}}$, $5^{\text{th}}$, and $6^{\text{th}}$ columns in this figure, we observe that these simple manipulations produce significant changes in image tone, and since color variation is a crucial aspect of style transfer, this finding further drives our focus on startpoint enhancement.

Third, by examining the results in the $3^{rd}$ and $5^{th}$ rows, we notice that the startpoint not only affects the tone of generated images but can also influence the content of generated images to some extent, such as the facial outline of the woman in the $3^{rd}$ row and the background in the $5^{th}$ row. This effect has been largely overlooked in previous works, yet it is undeniably critical for style transfer tasks.

In summary, through simple adjustments to the startpoint, we have discovered its substantial impact on style transfer results — affecting content preservation, content modification, and tonal changes. These insights have driven us to pursue sampling startpoint enhancement for style transfer research. Therefore, our method, StyleSSP, emphasizes guidance during the inversion step and manipulation of the inversion latent space to achieve a more effective sampling startpoint in style transfer issues.

\subsection{User Study}
\begin{figure}
	\centering
	\includegraphics[width=0.6\linewidth]{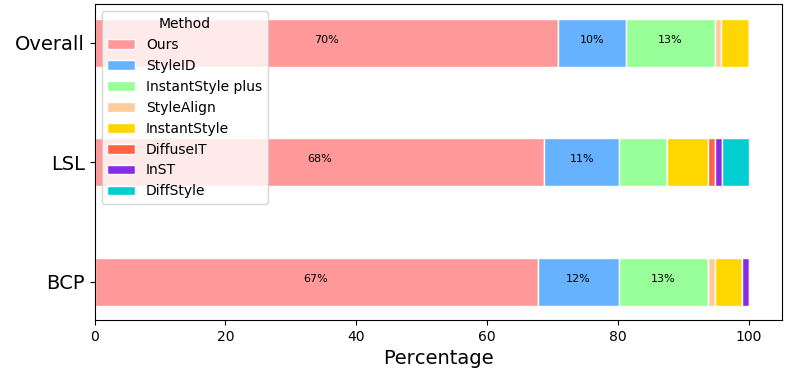}
	\caption{Results for the user study in percentages.}
	\label{fig:user study}
\end{figure}
We conduct added a user study based on the setting of Deadiff~\cite{deadiff}. We employed StyleID~\cite{StyleID}, StyleAlign~\cite{stylediffusion}, InstantStyle plus~\cite{InstantStyle-Plus}, InstantStyle~\cite{InstantStyle}, DiffuseIT~\cite{DiffuseIT}, DiffStyle~\cite{DiffStyle}. Additionally, InST~\cite{InST} and StyleSSP to separately generate 4 stylized images. As shown in Fig.~\ref{fig:user study}, 24 users from diverse backgrounds evaluate there images in terms of best content preservation (BCP), least style leakage (LSL), and overall performance (Overall). StyleSSP outperforms all state-of-the-art methods on three evaluation aspects with a big margin, which demonstractes the broad application prospects of our method.

\subsection{Parameter Selection}
\begin{figure*}
	\centering
	\includegraphics[width=0.8\linewidth]{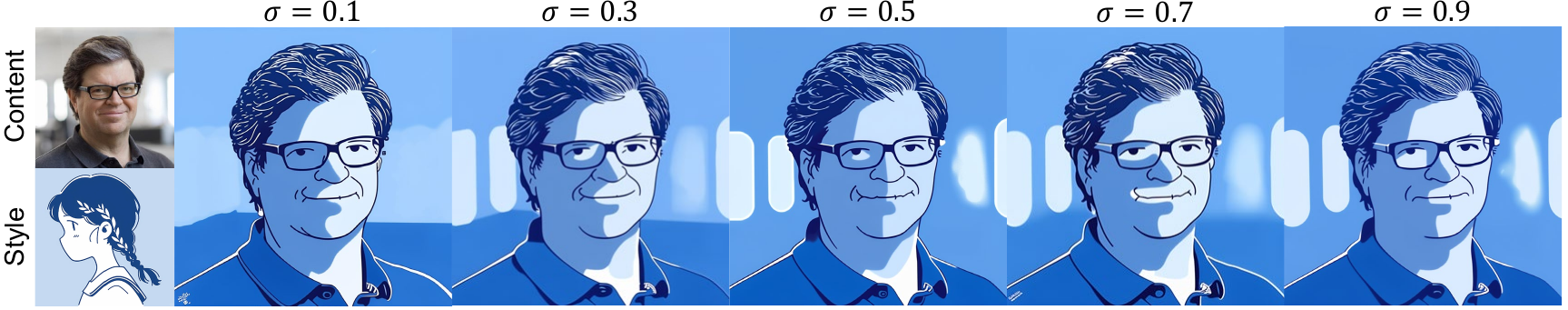}
	\caption{Visualization of frequency pass parameter $\sigma$'s effect.}
	\label{fig:R2}
\end{figure*}
\begin{figure}
	\centering
	\includegraphics[width=0.6\linewidth]{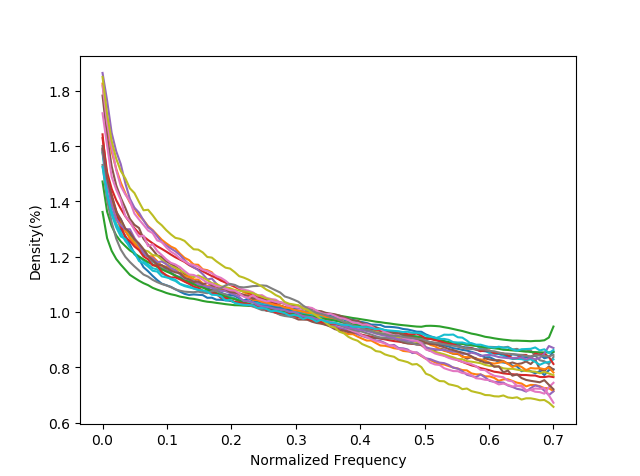}
	\caption{Frequency spectrum distribution of 20 random images.}
	\label{fig:image frequency}
\end{figure}
We conducted additional experiments to show how the frequency pass parameter $\sigma$ affects the results. As shown in Fig.~\ref{fig:R2}, $\sigma$ performs best in the range of 0.3 to 0.5, performing the best background and facial line preservation. This is because a very small $\sigma$ fails to emphasize high-frequency information, while a very large $\sigma$ suppresses too many valid components of images. Moreover, Fig.~\ref{fig:image frequency} shows that, although images differ in the spatial domain, their frequency distributions are quite similar, which supports us to use nearly the same $\sigma$ for different images. Moreover, since the frequency distribution of images is similar, the frequency band related to contours is not significantly affected by the choice of diffusion model. Therefore, different diffusion models should share the same frequency pass parameter $\sigma$. In summary, we recommend choosing $\sigma$ between $0.3$ and $0.5$, and this choice of value is not significantly related to the diffusion model.

\subsection{Principle of Negative Guidance}
\label{sec: principle of Negative Guidance}

\begin{figure*}[htbp]
	\centering
	\includegraphics[width=\linewidth]{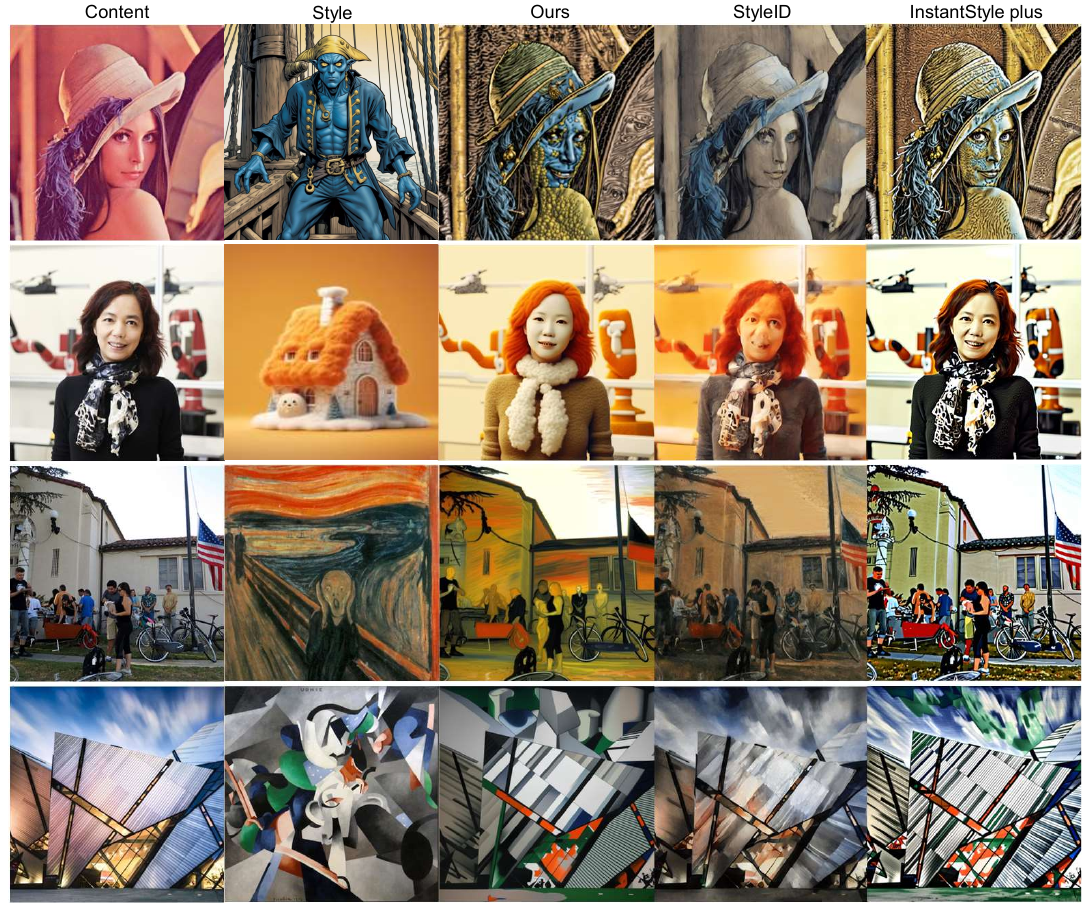}
	\caption{Qualitative comparison with with baselines(StyleID, InstantStyle plus). Zoom in for viewing details.}
	\label{fig:comparison_2}
\end{figure*}

In this section, we provide a detailed introduction to the principles of negative prompt guidance, starting with conditional generation. For conditional generation, that is, to sample samples from the conditional distribution $p(x|y)$. According to the Bayes formula, we can obtain: 

\begin{equation}
	\begin{aligned}
		p(x|y)&=\frac{p(y|x)p(x)}{p(y)},\\
		log~p(x|y)&=log~p(y|x)+log~p(x)-log~p(y),\\
		\Rightarrow \nabla_xlog~p(x|y)&=\nabla_xlog~p(y|x)+\nabla_xlog~p(x).
	\end{aligned}
\end{equation}

In the classifier-guided task, the score-based model with unconditional input is an estimation of $\nabla_x \log p(x)$, so in order to obtain $\nabla_x \log p(x|y)$, an additional classifier needs to be trained to estimate $\nabla_x \log p(y|x)$. At the same time, to control the strength of condition, the guidance scale $\omega$ is introduced:

\begin{equation}
	\nabla_xlog~p(x|y):=\omega\nabla_xlog p(y|x)+\nabla_xlog~p(x).
\end{equation}

In classify-free guidance (CFG) tasks, they simultaneously train two score-based models, $\nabla_x log~p(x)$ and $\nabla_x log~p(y|x)$. Since $\nabla_xlog~p(y|x)=\nabla_xlog~p(x|y)-\nabla_x log~p(x)$, it follows that: 

\begin{equation}
	\nabla_x log~p(x|y):=\omega(\nabla_xlog~p(x|y)-\nabla_x log~p(x))+\nabla_xlog~p(x),
\end{equation}

When negative prompt serves as a condition, the conditions for diffusion model contain two items, one is positive prompt condition $y$, and the other is negative prompt condition $\text{not}~\tilde{y}$. Since re-training a score-based model to estimate $\nabla_x p(x|y,\text{not}~\tilde{y})$ is costly, the following simplification is made:

\begin{equation}
	\label{14}
	\begin{aligned}
		p(x|y,\text{not}~\tilde{y})&=\frac{p(x,y,\text{not}~\tilde{y})}{p(y,\text{not}~\tilde{y})}\\
		&=\frac{p(y|x)p(\text{not}~\tilde{y}|x)p(x)}{p(y,\text{not}~\tilde{y})}\\
		&\propto\frac{p(x)}{p(y,\text{not}~\tilde{y})}\frac{p(y|x)}{p(\tilde{y}|x)},
	\end{aligned}
\end{equation}
so that:
\begin{equation}
	\label{15}
	\begin{aligned}
		\nabla_xlog~p(x|y,\text{not}~\tilde{y})&\propto \nabla_xlog~p(x)\\
		&+\nabla_x log~p(y|x)-\nabla_xlogp(\tilde{y}|x).
	\end{aligned}
\end{equation}

The Eq.~\ref{14} and Eq.~\ref{15} assume that $x$, $y$ and $\text{not}~\tilde{y}$ are mutually independent. Letting $\omega^+$ be the guidance scale of positive condition and $\omega^{-}$ be the guidance scale of negative condition, we have: 

\begin{equation}
	\begin{aligned}
		\nabla_xp(x|y,\text{not}~\tilde{y})&:=\nabla_xp(x)+\omega^+(\nabla_xp(x|y)-\nabla_xp(x))\\
		&-\omega^-(\nabla_xp(x|\tilde{y})-\nabla_xp(x)).
	\end{aligned}
\end{equation}

Thus, we can estimate $\nabla_xp(x|y,\text{not}~\tilde{y})$ only by calculating $\nabla_xp(x),\nabla_xp(x|y),\nabla_xp(x|\tilde{y})$, and all of these can be obtained through the pre-trained diffusion model.

It should be noted that in the negative guidance method proposed in this paper, IP-Instruct merely exists as a style and content extractor, which can be replaced by any other extractor. Meanwhile, this CFG-based guidance can also be replaced by the gradient-based guidance like FreeTune~\cite{FreeTuner} does. We emphasize that our prominent contribution lies in discovering that by guiding the startpoint of sampling stage to distance from the style image's content, thereby preventing the content leakage from style image.

\subsection{Additional Results}
\label{sec:More example}
We additionally compare the proposed method with the most recent baseline (StyleID) and the baseline with lowest ArtFID (InstantStyle plus). Fig.~\ref{fig:comparison_2} shows the additionally qualitative comparison of ours with diffusion model baselines. 

\begin{figure*}[htbp]
	\centering
	\includegraphics[width=\linewidth]{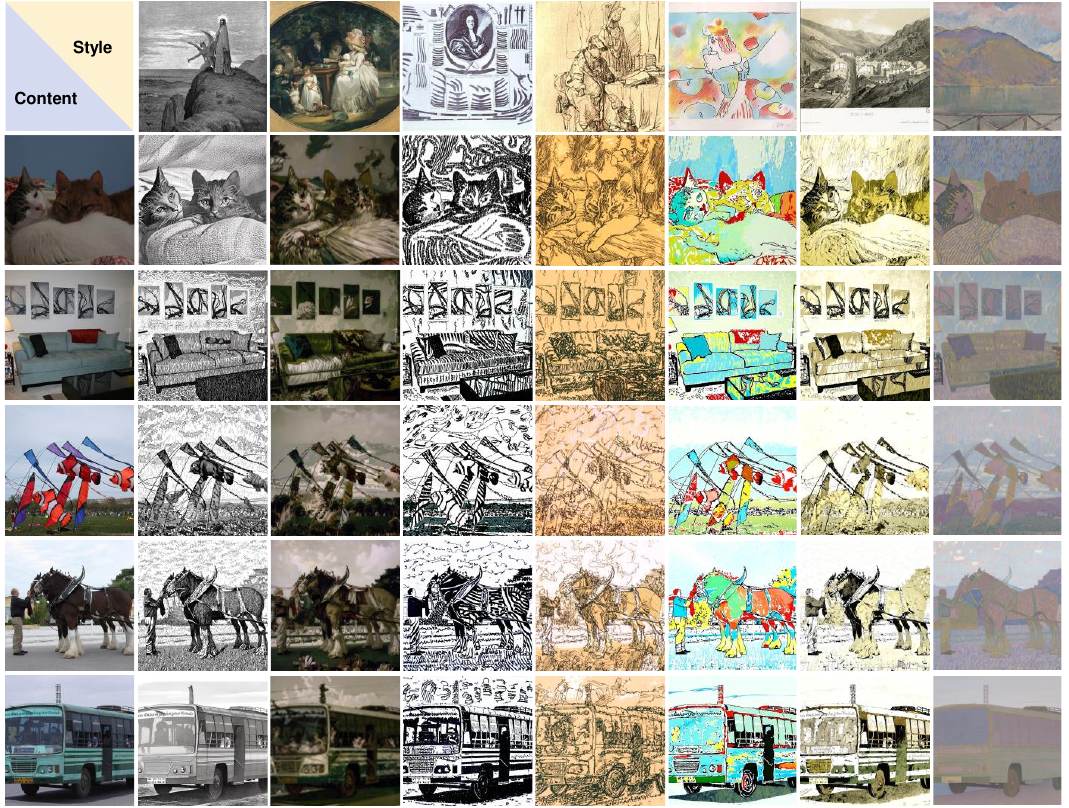}
	\caption{Style transfer results of style and content image pairs. Zoom in for viewing details.}
	\label{fig:ours_2}
\end{figure*}

Also, in Fig.~\ref{fig:ours_2}, we visualize the style transfer results of various pairs of content and style images, which further demonstrate StyleSSP's robustness and versatility in adapting to diverse content and style.

\end{document}